\documentclass[11pt]{article}

\usepackage[final]{acl}
\usepackage{times}
\usepackage{latexsym}

\usepackage[T1]{fontenc}

\usepackage[utf8]{inputenc}

\usepackage{microtype}

\usepackage{inconsolata}

\usepackage{graphicx}
\usepackage{enumitem}

\usepackage{hyperref}
\usepackage{url}
\usepackage{booktabs}
\usepackage{multirow}
\usepackage{geometry}
\usepackage[table]{xcolor}
\usepackage{lineno}
\usepackage{tikz}
\usetikzlibrary{positioning, arrows.meta, shapes.geometric, calc, decorations.pathmorphing}
\usepackage{caption}
\usepackage{subcaption}
\usepackage{stfloats}
\usepackage{amsmath}
\usepackage[T1]{fontenc}
\usepackage[utf8]{inputenc}
\usepackage{amssymb}
\usepackage[most]{tcolorbox}
\tcbuselibrary{skins, listings, breakable}

\usepackage{cuted}
\usepackage{placeins}
\definecolor{darkblue}{rgb}{0, 0, 0.5}
\definecolor{PromptBg}{HTML}{F2F2EF}
\definecolor{PromptBorder}{HTML}{4A4A4A}
\definecolor{PromptTitleBg}{HTML}{4A4A4A}

\hypersetup{colorlinks=true, citecolor=darkblue, linkcolor=darkblue, urlcolor=darkblue}

\newcommand{\ours}{\textsc{\textbf{MERIT}}}

\newtcblisting{promptbox}[2][]{%
  enhanced,
  breakable,        
  lines before break=1,
  listing only,
  width=\textwidth,   
  colback=PromptBg,
  colframe=PromptBorder,
  boxrule=0.9pt,
  arc=2mm,
  outer arc=2mm,
  left=8pt,
  right=8pt,
  top=8pt,
  bottom=8pt,
  colbacktitle=PromptTitleBg,
  coltitle=white,
  fonttitle=\bfseries\scshape\large,
  title={#2},
  listing options={
    basicstyle=\ttfamily\small,
    breaklines=true,
    columns=fullflexible,
    keepspaces=true,
    showstringspaces=false,
    upquote=true,
    literate=
      {→}{{$\rightarrow$}}1
      {✓}{{\checkmark}}1
      {✗}{{$\times$}}1
  },
  #1
}

\title{Learning to Retrieve: \\Dual-Level Long-Term Memory for Text-to-SQL Agents}

\author{
    Yibo Wang \\
    University of Illinois Chicago \\
    \texttt{ywang633@uic.edu} \\
    \And
    Nikki Lijing Kuang \\
    Snowflake AI Research \\
    \texttt{nikki.kuang@snowflake.com} \\
    \AND
    Philip S. Yu \\
    University of Illinois Chicago \\
    \texttt{psyu@uic.edu} \\
    \And
    Zhewei Yao \\
    Snowflake AI Research \\
    \texttt{zhewei.yao@snowflake.com} \\
    \And
    Yuxiong He \\
    Snowflake AI Research \\
    \texttt{yuxiong.he@snowflake.com}
}

\begin{document}
\maketitle

\begin{abstract}

Interactive text-to-SQL agents solve database tasks through multi-turn interactions involving schema exploration, query execution, feedback interpretation, and decision revision. Long-term memory helps agents reuse past experiences, but existing retrieval methods remain limited. 
Static methods use fixed similarity heuristics, whereas dynamic methods often learn from sparse terminal outcomes and retrieve at a single decision horizon.
This is insufficient when memory usefulness changes across interaction stages, since memories useful for initial planning may differ from those needed for local, state-conditioned execution.
We propose \ours{}, a dynamic multi-horizon memory retrieval framework. \ours{} maintains episode-level memory for global strategic guidance and turn-level memory for local decision support. Both levels use learned retrieval policies optimized with reinforcement learning. To train turn-level retrieval despite limited intermediate supervision, \ours{} uses a lightweight Process Reward Model to provide dense proxy rewards for local memory selection.
On BIRD-Interact, \ours{} achieves higher success rates and fewer interaction turns than no-memory, static-retrieval, and dynamic-retrieval baselines.
Transfer results on Spider2-Snow further show that the learned policies transfer without degradation, staying above all memory baselines without benchmark-specific tuning.
These results suggest that multi-horizon retrieval improves experience reuse in interactive text-to-SQL agents.
Our code is available at \url{https://github.com/YiboWANG214/MERIT}.

\end{abstract}

\section{Introduction}

\begin{figure*}[t]
    \centering

    \begin{subfigure}[t]{0.4\textwidth}
        \centering
        \includegraphics[width=\linewidth]{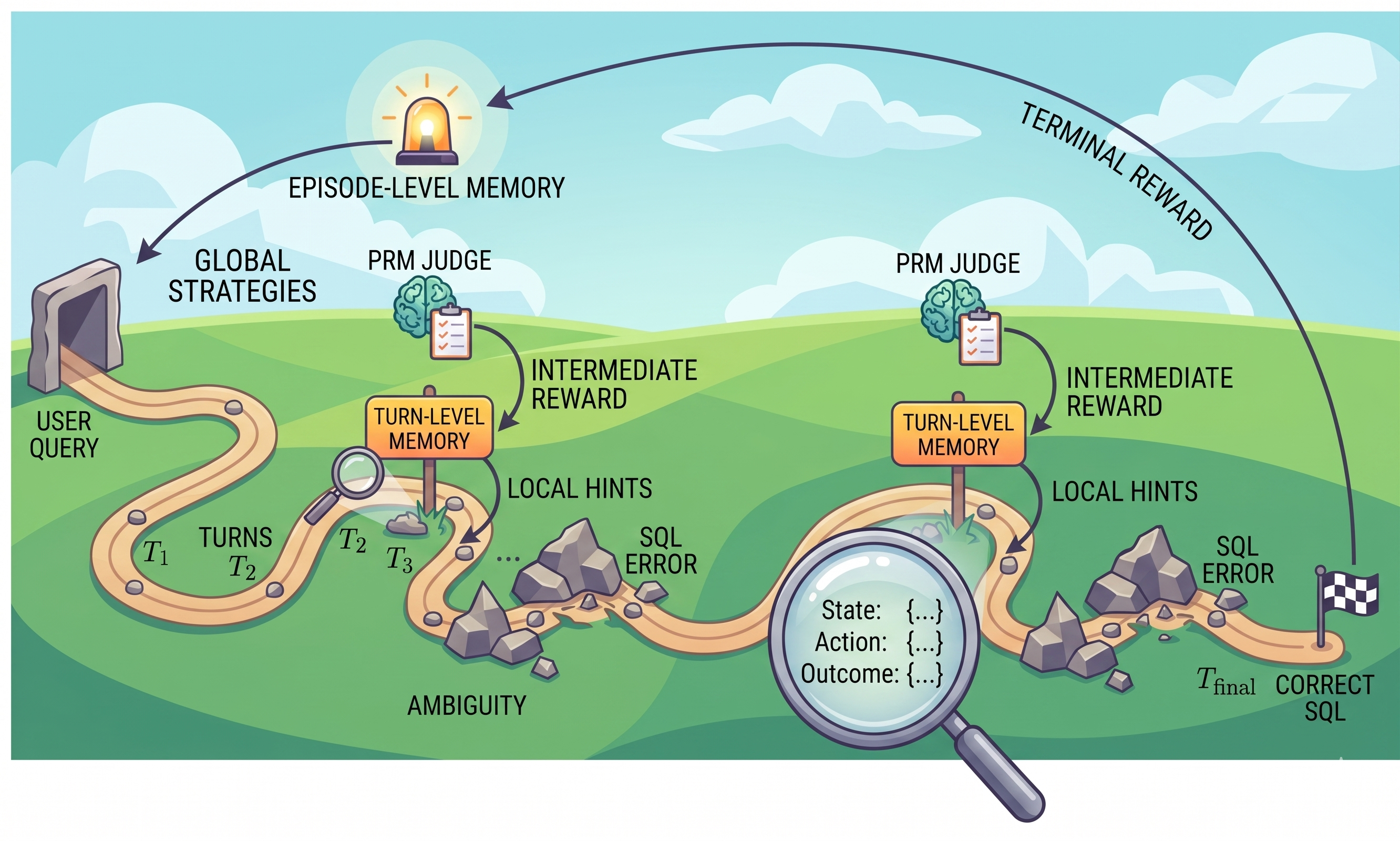}
        \caption{Conceptual overview of \ours{}.\protect\footnotemark}
        \label{fig:overview_a}
    \end{subfigure}\hfill
    \begin{subfigure}[t]{0.58\textwidth}
        \centering
        \includegraphics[width=\linewidth,trim=6 4 6 4,clip]{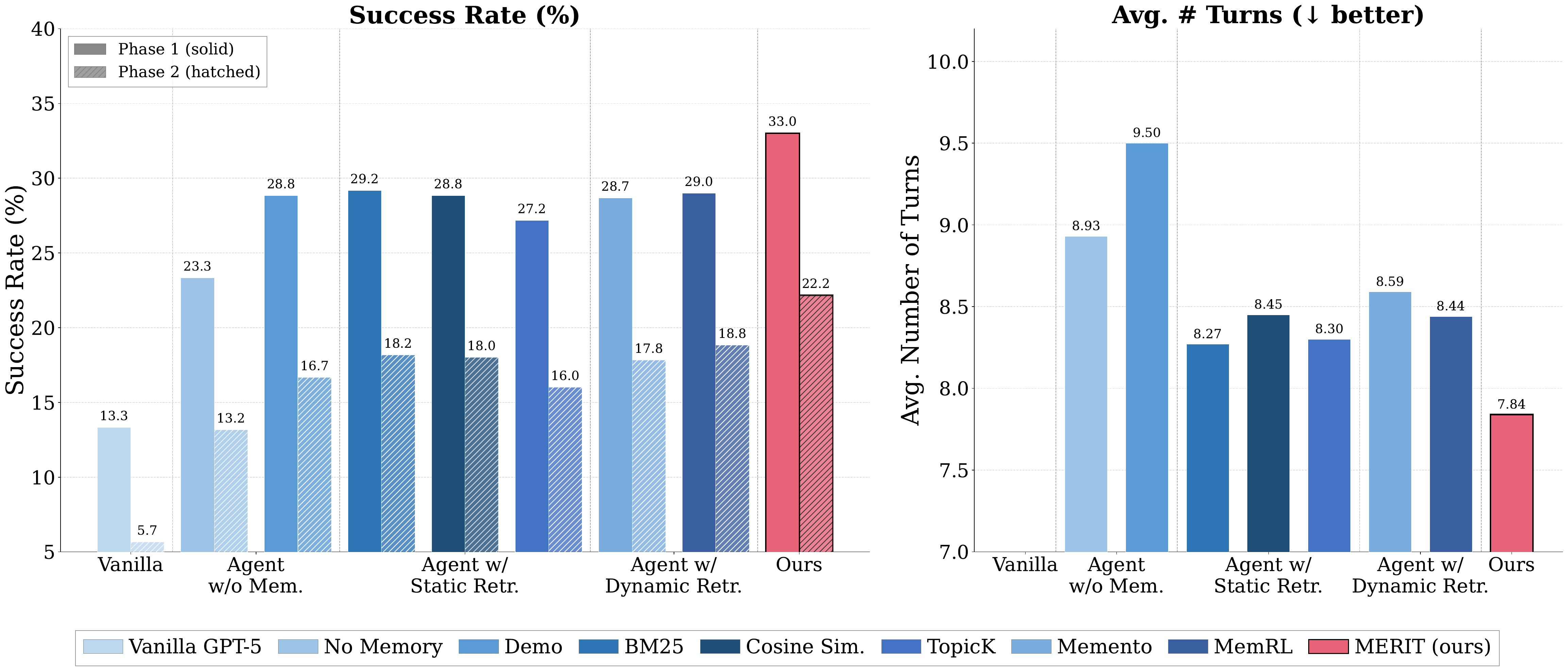}
        \caption{Performance comparison with baselines.}
        \label{fig:overview_b}
    \end{subfigure}

    \caption{\textbf{Conceptual Overview and Performance Advantages of \ours{}.}
    (a) \ours{} performs horizon-aware retrieval by decoupling memory into episode-level (global strategy) and turn-level (local hints), using a PRM to provide the dense rewards required for turn-level policy learning.
    (b) \ours{} outperforms static and single-horizon dynamic baselines in success rates while requiring fewer interaction turns.}
    \label{fig:main_overview}
\end{figure*}

Large language models (LLMs) are increasingly deployed as multi-step interactive agents that act, observe, and refine their behavior to solve complex tasks~\citep{li2025review,mohammadi2025evaluation,xu2025llm,yuan2025easytool}. In text-to-SQL, this shift moves the task beyond single-pass semantic parsing. Instead of producing SQL in a single step, an agent may need to inspect schemas, execute candidate SQL queries, ask clarifying questions, observe feedback, and revise its plan over time.
Thus, success depends not only on generating a valid final query but also on making reliable intermediate decisions under evolving context~\citep{wang-etal-2025-mac,shao2025enhancing}. 

Long-term memory provides a natural mechanism for improving such decisions. Interactive text-to-SQL tasks often share recurring structures and solution patterns, including join templates, aggregation logic, clarification strategies, and debugging routines. Reusing these experiences can help agents avoid redundant exploration and make informed decisions in new interactions.
However, existing memory-augmented approaches commonly rely on static similarity heuristics~\citep{maharana2024evaluating,sun2026h,salama2025meminsight,yan2025memory-r1,kweon2025topick}, such as dense embedding similarity, lexical matching, or hand-designed schema heuristics. These heuristics are useful for finding related candidates, yet do not directly optimize whether a retrieved memory will improve the next decision.

Recent dynamic frameworks address this limitation by learning memory selection from downstream feedback. For example, Memento~\citep{zhou2025memento} and MemRL~\citep{zhang2026memrl} update retrieval policies or memory utilities based on final outcomes.
However, these methods mainly retrieve at the episode level, selecting full trajectories based on the initial request and learning from sparse terminal rewards.
This mismatches interactive text-to-SQL, where the agent's state evolves through schema exploration, query execution, observations, and plan revision. Thus, retrieval should match the decision horizon.
For example, a trajectory useful for initial planning may not help after a SQL error, where a local debugging pattern is more relevant.
Moreover, a final success or failure signal provides weak supervision for determining which memory was useful at a specific intermediate turn.

Therefore, we propose a dynamic \textbf{M}ulti-horizon m\textbf{E}mory \textbf{R}etrieval framework using \textbf{I}ntermediate and \textbf{T}erminal rewards (\ours{}). \ours{} formulates memory retrieval as a learning problem across two decision horizons.
As illustrated in Figure~\ref{fig:overview_a}, the episode-level retriever selects past trajectories before an interaction begins, providing global strategic guidance for planning. The turn-level retriever selects compact state-action-observation memories during the interaction, providing local hints conditioned on the current state. This dual-level design separates long-range task strategy from short-range execution support.

{\footnotetext{Figure generated by \href{https://gemini.google.com/app}{Gemini}.}}
\ours{} uses a two-stage retrieval pipeline at both levels. An embedding retriever first generates a candidate set, and a learned selector then chooses which memories to expose to the agent. Both selectors are optimized with reinforcement learning (RL), shaping retrieval by downstream task utility rather than fixed similarity alone. 
The episode-level selector is trained with terminal rewards since its retrieval decision affects the overall trajectory. 
The turn-level selector requires denser supervision because intermediate turns do not have reliable direct outcome labels~\citep{choudhury2025process,xi2025agentprm}. 
To address this credit-assignment challenge, \ours{} introduces a lightweight Process Reward Model (PRM) that evaluates the utility of retrieved turn-level memories with a structured rubric. The resulting dense rewards enable RL for state-conditioned turn-level retrieval.

We evaluate \ours{} on the interactive text-to-SQL benchmark~\citep{huo2025bird} for in-domain interactive performance and on Spider2-Snow~\citep{lei2025spider} for cross-benchmark transfer. As summarized in Figure~\ref{fig:overview_b}, \ours{}
achieves higher success rates than no-memory, static-retrieval, and dynamic-retrieval baselines, while reducing the average number of interaction turns in the in-domain setting.
On Spider2-Snow, \ours{} transfers without benchmark-specific training or tuning and stays above all baselines, suggesting that the learned retrieval policies do not overfit the source benchmark.

Our contributions are summarized as follows:
\begin{itemize}
[noitemsep,topsep=0.6pt,leftmargin=0.5cm]
    \item We propose \ours{}, which aligns memory retrieval with decision horizon through episode-level global guidance and turn-level state-conditioned local support.
    \item We formulate memory retrieval as learned policies at both decision horizons, allowing memory selection to be optimized by downstream task utility rather than fixed similarity alone.
    \item We introduce a lightweight PRM that provides dense structured rewards for turn-level retrieval, mitigating the credit-assignment problem caused by sparse terminal outcomes.
    \item We evaluate \ours{} in both in-domain and transfer settings, showing improved success rates and interaction efficiency in the in-domain setting, and cross-benchmark transfer without degradation.
\end{itemize}

\section{Related Work}

\subsection{Text-to-SQL}
Early neural text-to-SQL research typically formulates the task as single-pass semantic parsing from a natural-language question and database schema directly to SQL~\citep{zhong2017seq2sql,xu2017sqlnet,yu-etal-2018-typesql,lin-etal-2020-bridging}. Recent LLM-based approaches move beyond one-shot generation by introducing interaction, tool use, and iterative refinement~\citep{zhang-etal-2024-coe, tian-etal-2023-interactive, xiong2024interactive, wang-etal-2025-mac, zhu2026residual}. 
These studies show that text-to-SQL agents benefit from intermediate feedback and adaptive decision-making. We build on this interactive setting and study experience reuse through learned long-term memory retrieval.

\subsection{Memory Systems in LLM Agents}
Long-term memory enables cross-episode experience reuse~\citep{park2023generative, lu2023memochat}, and prior work has explored long-term memory writing and organization, including operating-system-like memory management~\citep{packer2023memgpt}, semantic compression~\citep{liu2026simplemem}, graph-based memory representations ~\citep{chhikara2025mem0}, trajectory distillation~\citep{fang2025memp}, and structural memory representations~\citep{zeng2024structural}. These studies improve the storage and representation of agent experience.
\ours{} instead focuses on retrieving stored memories in the interactive setting.
Most memory-augmented agents retrieve memories with fixed heuristics, such as dense embedding similarity, lexical matching, topic coverage, or predefined structural relations~\citep{arslan2024survey, kweon2025topick, kagaya2024rap, zeng2024structural}. 
Even systems that learn other memory operations often keep retrieval fixed; for example, Memory-R1~\citep{yan2025memory-r1} trains a memory manager and an answer agent with reinforcement learning but still retrieves through similarity-based RAG. 
These methods are effective for finding related candidates, yet their retrieval criteria are fixed and do not directly optimize downstream utility.
Recent dynamic methods learn memory utility from feedback. 
MemRL~\citep{zhang2026memrl} updates memory Q-values through runtime learning, and Memento~\citep{zhou2025memento} learns a memory-selection policy without fine-tuning base LLMs. 
They move beyond static similarity, but primarily operate at the episode level and rely on sparse terminal feedback. 
Learned retrieval beyond terminal reward has also been studied for test-time search, such as \citet{ICLR2024_25f7be96}, who train a model to adaptively decide when and what to retrieve at inference time, though their retrieval target is external documents at a single horizon. 
\ours{} instead retrieves the agent's own memory, separating episode-level retrieval for global strategy from turn-level retrieval for local support, and trains the turn-level retriever with dense PRM-based rewards.

\section{Preliminaries}

\label{sec:setting}

We formulate interactive text-to-SQL as a sequential decision-making process. 
Given an initial natural-language request $x$, the agent interacts with a database environment and, when necessary, a user simulator. Unlike single-pass semantic parsing, the agent may take multiple actions before producing the final SQL query, including schema inspection, knowledge retrieval, SQL execution, clarification, and final submission.

At turn $t$, the agent has an interaction history
\[
h_{t-1} = \big((a_1,o_1), (a_2,o_2), \ldots, (a_{t-1},o_{t-1})\big),
\]
where $a_t$ is the agent action, and $o_t$ is the corresponding observation. The agent observes a state
\[
s_t = (x, h_{t-1}, b_t),
\]
where $b_t$ is the remaining interaction budget. Conditioned on $s_t$, the agent selects an action $a_t$, receives an observation $o_t$, and appends $(a_t,o_t)$ to the history.
An episode yields a trajectory
\[
\tau = (x, h_T),
\]
where $T$ is the terminal turn.
The episode terminates when the agent submits a final SQL query, exhausts its interaction budget, or reaches the maximum allowable number of turns.
The final query is evaluated by execution correctness against the target SQL semantics.

\section{Data Augmentation}

\label{sec:data_aug}

Learning retrieval policies requires diverse executable SQL. Since high-quality interactive text-to-SQL data is limited, we construct additional training data through an execution-grounded augmentation pipeline. 
Rather than perturbing natural-language requests directly, we first synthesize executable SQL targets from seed examples and then generate user requests grounded in those targets. This SQL-first design reduces semantic drift and encourages executable SQL, faithful SQL-query alignment, and structural diversity.

Given a seed query, we apply constrained SQL-level perturbations, including compatible column substitutions, numeric literal perturbations, sorting-direction changes, and limit-value modifications. These operations diversify projections, grouping, thresholds, rankings, and result sizes while preserving executability. For each augmented SQL target, an LLM generates a fully specified request, an intentionally underspecified request, and grounding metadata that links underspecified phrases to corresponding SQL fragments. We discard examples whose generated text or grounding metadata cannot be verified against the SQL logic. Each retained example therefore provides an executable text-to-SQL task with aligned SQL, requests, and grounding metadata for retrieval-policy training. 
Additional details, prompts, and a diversity and quality analysis of the augmented data are provided in Appendix~\ref{app:data_aug}.

\section{Methodology}
\begin{figure*}[t]
\begin{center}
\includegraphics[width=\linewidth]{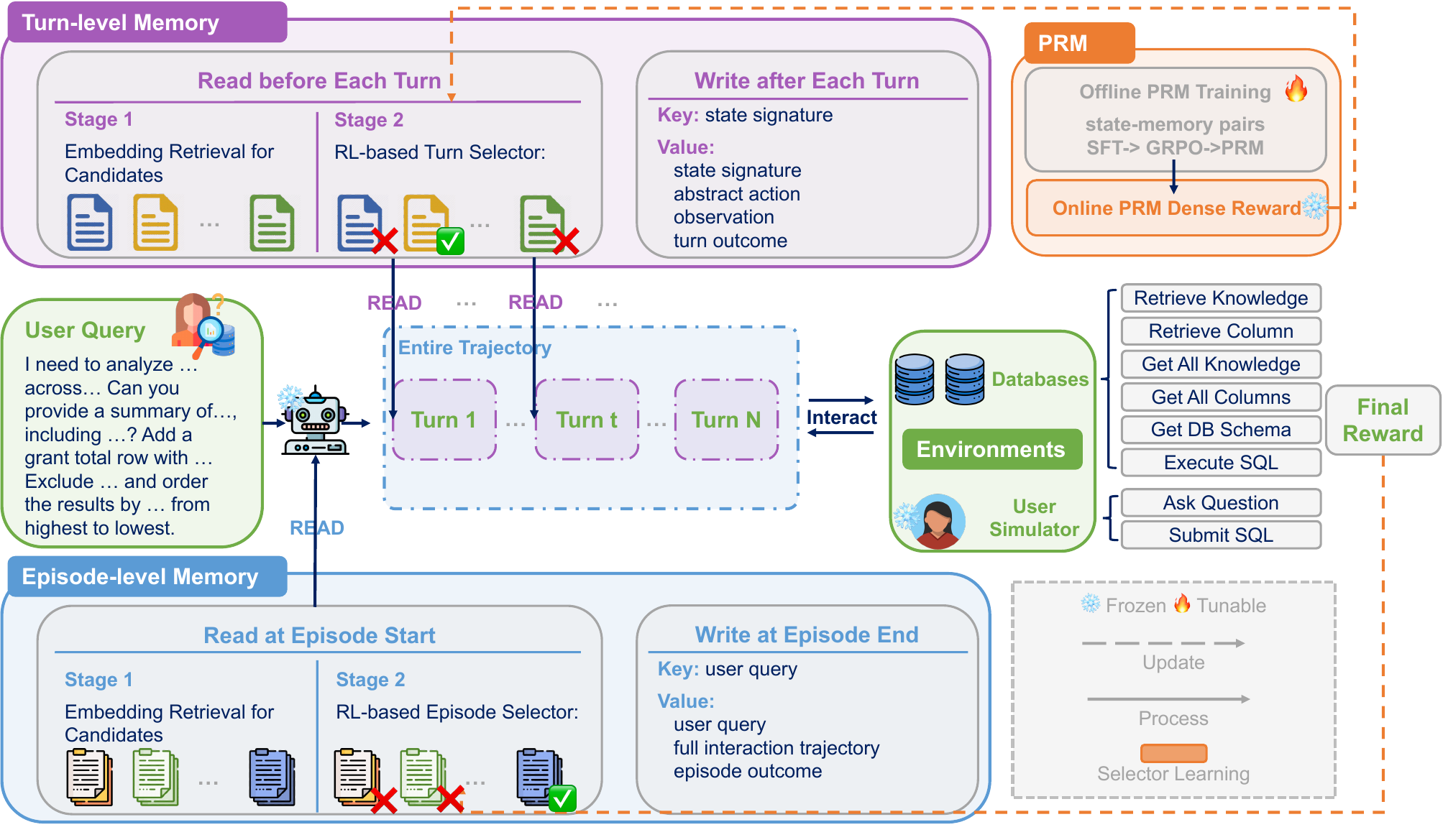}
\end{center}
\caption{Overview of the proposed dual-level long-term memory framework for interactive Text-to-SQL. Episode-level memory is retrieved at episode start for global strategic guidance and updated at episode end, while turn-level memory is retrieved and updated per turn to provide state-conditioned local hints. Both levels employ a two-stage pipeline: embedding-based candidate generation followed by an RL-trained selector. A Process Reward Model (PRM), trained offline with SFT and GRPO, provides dense online rewards for turn-level retrieval, while episode-level retrieval is optimized with terminal task reward.}
\label{fig: model_framework}
\end{figure*}


We present \ours{} by first introducing the dual-level memory architecture that separates global strategic guidance from local state-conditioned support (\S~\ref{sec:memory}). We then formulate memory retrieval at both levels as learned policies optimized by reinforcement learning (\S~\ref{sec:retrieval}). Finally, we describe the Process Reward Model (PRM), which provides dense supervision for turn-level retrieval (\S~\ref{sec:prm}). The complete architecture is shown in Figure~\ref{fig: model_framework}.

\subsection{Dual-Level Long-Term Memory}
\label{sec:memory}
\ours{} maintains two long-term memory banks. Episode-level memory stores complete interactions, and turn-level memory stores compact state-action-observation snippets for local decisions.

\paragraph{Episode-level memory.}
After each completed episode $i$, we write an episode-level memory
\[
m_i^{E} = (k_i^{E}, v_i^{E}),
\]
where $k_i^{E}$ is the embedding of user request $x_i$ and
\[
v_i^{E} = (x_i, \tau_i, \ell_i, z_i, g_i)
\]
stores the original request $x_i$, the full trajectory $\tau_i$, the terminal outcome label $\ell_i$, distilled episode insights $z_i$, and structured task-specific constraints or facts $g_i$ resolved during interaction. 
The distilled insights summarize reusable strategic knowledge, such as successful plans, common failure modes, task characteristics, and guidance for similar future tasks. The prompt for insights extraction is in Appendix~\ref{prompts: episode insight extraction prompt}.
 
For a new request $x$, we retrieve episode-level candidates by cosine similarity over the stored episode keys $\{k_i^{E}\}$. 
Conditioned on $x$ and compact candidate summaries, the episode-level selector chooses a subset of memories to inject into the initial agent prompt.
These memories provide global guidance before the interaction begins.

\paragraph{Turn-level memory.}
Turn-level memory is indexed by a compact state signature rather than the full interaction state.
At turn $t$, we define the state signature 
\[
\sigma_t = g(s_t),
\]
where $g(\cdot)$ maps the full state $s_t$ into a textual retrieval key.
The signature contains the original request, the current query context, a coarse description of the previous action, and the previous observation. We keep the previous action coarse in the retrieval key so that similar decision states can match even when their exact SQL queries differ.

After the agent completes turn $j$, we write a turn memory
\[
m_j^{T} = (k_j^{T}, v_j^{T}),
\]
where $k_j^{T}$ is the embedding of $\sigma_j$, and
\[
v_j^{T} = (\sigma_j, \tilde{a}_j, \tilde{o}_j, \ell_j^{T})
\]
stores the state signature, an abstracted action representation $\tilde{a}_j$, a short observation snippet $\tilde{o}_j$, and an execution-status label $\ell_j^{T}$. The label records immediate action outcomes, such as invalid tool calls and SQL syntax errors, but does not capture whether it contributes to broader task progress.
The abstracted action may include normalized SQL skeletons or other tool-call details, allowing reusable local patterns to transfer across queries.

Before intermediate turn $t$, we retrieve turn-level candidates by cosine similarity over $\{k_j^{T}\}$.
Conditioned on $\sigma_t$ and compact candidate summaries, the turn selector returns a subset of memories to inject as brief hints for the next agent action.

\subsection{Retrieval Policy Learning}
\label{sec:retrieval}
\ours{} learns retrieval policies that decide which candidate memories to expose to the agent. At both levels, 
similarity-based retrieval first obtains candidates, and then a learned selector chooses a subset for prompt injection.
We optimize both selectors with Proximal Policy Optimization (PPO), using reward signals aligned with their respective decision horizons.

\paragraph{Episode-level retrieval policy learning.}
Given an initial request $x$ and an episode-level candidate set $C^{E}(x)$, the episode-level selector samples a retrieval decision
\[
d^{E} \sim \pi^{E}_{\theta}(\cdot \mid x, C^{E}(x)),
\]
where $d^{E}$ specifies the selected memory subset $\hat{M}^E \subseteq C^E(x)$. The selected memories are injected before the interaction begins.
After the agent completes trajectory $\tau$, the selector receives a terminal reward $R^{E}(\tau)$, derived from final execution correctness.
We optimize $\pi_\theta^E$ with PPO to learn episode-level retrieval from downstream outcomes.

\paragraph{Turn-level retrieval policy learning.}
Given a state signature $\sigma_t$ and a turn-level candidate set $C^{T}(\sigma_t)$, the turn-level selector samples
\[
d_t^{T} \sim \pi^{T}_{\phi}(\cdot \mid \sigma_t, C^{T}(\sigma_t)),
\]
where $d_t^{T}$ specifies the selected memory subset $\hat{M}_t^T \subseteq C^T(\sigma_t)$. The selected memories are injected as local hints before the next agent action.
Turn-level retrieval cannot be trained directly from reliable immediate environment rewards. A memory selected at an intermediate turn may affect later actions, while its contribution may only be reflected in the final episode outcome. To mitigate this credit-assignment challenge, we use the PRM to estimate the utility of selected turn-level memories. The PRM scores each selected state-memory pair, and the turn-level reward is computed as
\[
r_t^T =
\frac{1}{|\hat{M}_t^T|}
\sum_{m \in \hat{M}_t^T}
f_{\mathrm{PRM}}(\sigma_t, m),
\]
where $f_{\mathrm{PRM}}$ maps a state signature and a memory to a scalar utility score. We optimize $\pi_\phi^T$ with PPO using these dense rewards.

\subsection{Process Reward Model for Turn-Level Supervision}
\label{sec:prm}

While turn-level retrieval requires dense supervision, terminal rewards provide delayed, noisy feedback on intermediate choices. \ours{} therefore uses a Process Reward Model (PRM) to estimate candidate memory utility given the current state.

\paragraph{Structured Utility Prediction.}
Given a state-memory pair $(\sigma_t, m)$, the PRM predicts a structured rubric
\[
y_t = f_{\psi}(\sigma_t, m).
\]
The rubric evaluates memory usefulness along seven dimensions: state match, actionability value, pattern generalizability, outcome reliability, clarity for agent, confidence in assessment, and overall utility. Each dimension contains a score in $[0,5]$ and a brief rationale. The rubric also includes binary decisions on whether to inject the memory and whether to use it as a warning, along with concise textual feedback on how to use it. The full rubric schema and prompt are provided in Appendix~\ref{prompts: prm judging prompt}.



For turn-level retrieval policy optimization, we use the trained PRM to score each state-memory pair. We denote the resulting scalar utility score as $f_{\mathrm{PRM}}(\sigma_t, m)$, which is extracted from the predicted rubric using the overall utility score and the injection decision. This score serves as the dense reward for turn-level PPO, while the structured rubric grounds the reward in interpretable aspects of memory usefulness.

\paragraph{Two-Stage Training.}
We train the PRM from teacher-annotated state-memory pairs. For each pair, a stronger teacher model produces the full rubric. We first apply supervised fine-tuning (SFT) to imitate the teacher rubrics, then refine the PRM with GRPO using a structured matching reward.

Let $\hat{y}$ denote the PRM-generated rubric and $y^{*}$ denote the teacher rubric. If $\hat{y}$ is not a valid JSON object with all required fields, we assign
\[
R_{\mathrm{PRM}}(\hat{y}, y^{*}) = -1.
\]
Otherwise, we first compute a structured matching score $S_{\mathrm{PRM}}$, and then clip it to obtain the PRM training reward:
\[
S_{\mathrm{PRM}}
= 0.7 R_{\mathrm{dim}} + 0.2 R_{\mathrm{bin}}
+ 0.1 R_{\mathrm{text}} - 0.5 \mathbb{I}_{\mathrm{extra}},
\]
\[
R_{\mathrm{PRM}}(\hat{y}, y^{*})
= \mathrm{clip}(S_{\mathrm{PRM}}, -1, 1),
\]
where $\mathbb{I}_{\mathrm{extra}}=1$ if the output contains text outside the JSON object, and $0$ otherwise.

The dimension-level reward compares the seven scored rubric dimensions. For each dimension $d$, let $\hat{s}_d, s_d^{*} \in [0,5]$ be the predicted and teacher numeric scores, and let $\hat{r}_d, r_d^{*}$ be the corresponding short reasons. We define
\begin{equation}
\begin{aligned}
R_{\mathrm{dim}} = \frac{1}{|\mathcal{D}|} \sum_{d \in \mathcal{D}} \Bigg[ & 0.7 \cdot \max\!\left(0, 1 - \frac{|\hat{s}_d - s_d^{*}|}{5}\right) \\
& + 0.3 \cdot J(\hat{r}_d, r_d^{*}) \Bigg], \notag
\end{aligned}
\end{equation}
where $\mathcal{D}$ is the set of the seven rubric dimensions and $J(\cdot,\cdot)$ denotes Jaccard similarity between token sets. 
The binary term $R_{\mathrm{bin}}$ is the average exact-match accuracy over the two binary decisions, 
and $R_{\mathrm{text}}$ is the average similarity over the two free-text feedback fields. This reward encourages the PRM 
to produce valid structured outputs that match teacher utility scores, decisions, and rationales.

\section{Experiments}

\begin{table*}[t]
\centering
\resizebox{\linewidth}{!}{%

\newcommand{\g}[1]{$_{\textcolor{green!60!black}{#1}}$}
\newcommand{\rc}[1]{$_{\textcolor{red}{#1}}$}

\begin{tabular}{l lll lll lll}
\toprule
\multirow{2}{*}{\textbf{Methods}} & \multicolumn{3}{c}{\textbf{Phase 1}} & \multicolumn{3}{c}{\textbf{Phase 2}} & \multicolumn{3}{c}{\textbf{Avg. \#Turns} $\downarrow$} \\
\cmidrule(lr){2-4} \cmidrule(lr){5-7} \cmidrule(lr){8-10}
& \textbf{WRITE} & \textbf{READ} & \textbf{Overall} & \textbf{WRITE} & \textbf{READ} & \textbf{Overall} & \textbf{WRITE} & \textbf{READ} & \textbf{Overall} \\
\midrule
\rowcolor{red!15} \multicolumn{10}{c}{\textit{No Memory}} \\
\midrule
Vanilla GPT-5 & 24.71\rc{-18.97} & 8.69\rc{-6.33} & 13.33\rc{-10.00} & 12.07\rc{-8.04} & 3.05\rc{-7.28} & 5.67\rc{-7.50} & - & - & - \\
No Memory & 43.68 & 15.02 & 23.33 & 20.11 & 10.33 & 13.17 & 8.24 & 9.21 & 8.93 \\
Demo & \textbf{54.60}\g{+10.92} & 18.31\g{+3.29} & 28.83\g{+5.50} & 28.16\g{+8.05} & 11.97\g{+1.64} & 16.67\g{+3.50} & 8.44\rc{+0.20} & 9.93\rc{+0.72} & 9.50\rc{+0.57} \\
\midrule
\rowcolor{yellow!15} \multicolumn{10}{c}{\textit{Static Memory}} \\
\midrule
BM25 & 52.87\g{+9.19} & 19.48\g{+4.46} & 29.17\g{+5.84} & 28.74\g{+8.63} & 13.85\g{+3.52} & 18.17\g{+5.00} & 7.65\g{-0.59} & 8.53\g{-0.68} & 8.27\g{-0.66} \\
Cosine Similarity & 51.72\g{+8.04} & 19.48\g{+4.46} & 28.83\g{+5.50} & 28.74\g{+8.63} & 13.62\g{+3.29} & 18.00\g{+4.83} & 7.63\g{-0.61} & 8.79\g{-0.42} & 8.45\g{-0.48} \\
TopicK & 48.85\g{+5.17} & 18.31\g{+3.29} & 27.17\g{+3.84} & 25.86\g{+5.75} & 11.97\g{+1.64} & 16.00\g{+2.83} & 7.46\g{-0.78} & 8.64\g{-0.57} & 8.30\g{-0.63} \\
\midrule
\rowcolor{blue!15} \multicolumn{10}{c}{\textit{Dynamic Memory}} \\
\midrule
Memento & 50.57\g{+6.89} & 19.72\g{+4.70} & 28.67\g{+5.34} & 28.74\g{+8.63} & 13.38\g{+3.05} & 17.83\g{+4.66} & 7.80\g{-0.44} & 8.91\g{-0.30} & 8.59\g{-0.34} \\
MemRL & 51.15\g{+7.47} & 19.95\g{+4.93} & 29.00\g{+5.67} & 30.46\g{+10.35} & 14.08\g{+3.75} & 18.83\g{+5.66} & 7.58\g{-0.66} & 8.79\g{-0.42} & 8.44\g{-0.49} \\
\rowcolor{gray!15} \ours{} (ours) & 53.49\g{+9.81} & \textbf{24.64}\g{+9.62} & \textbf{33.01}\g{+9.68} & \textbf{34.30}\g{+14.19} & \textbf{17.22}\g{+6.89} & \textbf{22.17}\g{+9.00} & \textbf{7.09}\g{-1.15} & \textbf{8.15}\g{-1.06} & \textbf{7.84}\g{-1.09} \\
\bottomrule
\end{tabular}
}
\caption{Performance comparison of \ours{} against baselines on Success Rates across two phases and Average Number of Turns. Performance is partitioned into queries with WRITE operations and READ-only operations. The backbone of the main agent is GPT-5 and the user simulator is GPT-4o for all methods. Subscripts denote absolute differences relative to the No Memory baseline.}
\label{tab: main results}
\end{table*}

\begin{table}[t]
\centering
\resizebox{\linewidth}{!}{%

\newcommand{\g}[1]{$_{\textcolor{green!60!black}{#1}}$}
\newcommand{\rc}[1]{$_{\textcolor{red}{#1}}$}

\begin{tabular}{l l l}
\toprule
\textbf{Methods} & \textbf{Success Rate} & \textbf{Avg. \#Turns}\\
\midrule
\rowcolor{red!15} \multicolumn{3}{c}{\textit{No Memory}} \\
\midrule
Vanilla GPT-5 & 37.29\rc{-28.71} & - \\
No Memory & 66.00 & \textbf{13.29} \\
\midrule
\rowcolor{yellow!15} \multicolumn{3}{c}{\textit{Static Memory}} \\
\midrule
BM25 & 67.09\g{+1.09} & 14.58\rc{+1.29} \\
Cosine Similarity & 68.19\g{+2.19} & 13.48\rc{+0.19}\\
TopicK & 66.73\g{+0.73} & 14.54\rc{+1.25} \\
\midrule
\rowcolor{blue!15} \multicolumn{3}{c}{\textit{Dynamic Memory}} \\
\midrule
Memento & \underline{68.56}\g{+2.56} & 14.33\rc{+1.04}\\
MemRL & 66.91\g{+0.91} & 14.95\rc{+1.66} \\
\rowcolor{gray!15} \ours{} (ours) & \textbf{69.29}\g{+3.29} & \underline{13.44}\rc{+0.15} \\
\bottomrule
\end{tabular}
}
\caption{Transferability comparison on Spider2-Snow, with the main-agent backbone as GPT-5. TopicK, Memento, MemRL, and \ours{} directly apply the trained checkpoints or artifacts. Subscripts denote absolute differences relative to the No Memory baseline.}
\label{tab: spider results}
\end{table}

\subsection{Experimental Settings}

\paragraph{Datasets and Metrics.}
We evaluate \ours{} on BIRD-Interact~\citep{huo2025bird} for in-domain interactive text-to-SQL and on Spider2-Snow~\citep{lei2025spider} for cross-benchmark transfer. For BIRD-Interact, we train the retrieval policies and the PRM on BIRD-Interact-Lite and evaluate on the unseen BIRD-Interact-Full dataset. Since BIRD-Interact-Lite contains only 236 base tasks (excluding 64 overlapped examples with BIRD-Interact-Full), we apply the data augmentation pipeline in \S~\ref{sec:data_aug} and Appendix~\ref{app:data_aug} to construct approximately 7,000 training examples, with benchmark-specific conversion details in Appendix~\ref{appendix: experimental settings}. 

BIRD-Interact reports results in two benchmark-defined phases. Phase 1 evaluates the initial user request, while Phase 2 evaluates state-dependent follow-up requests. 
We measure success by execution correctness and report the average number of interaction turns as an efficiency metric. We also report READ and WRITE results separately, where READ tasks answer analytical queries without modifying the database, while WRITE tasks require state-changing operations. This split tests whether \ours{} helps different interaction types that may require different strategies. For Spider2-Snow, all retrieval artifacts and learned policies are trained only on BIRD-Interact and transferred without Spider2-Snow-specific training or tuning.

\paragraph{Baselines.}
We compare \ours{} with no-memory, static-retrieval, and dynamic-retrieval baselines. The no-memory baselines include Vanilla GPT-5, a non-agentic LLM baseline, No Memory Agent, the base interactive agent without historical trajectories, and Demo Agent, which uses the officially released static demonstrations (available only for BIRD-Interact). Static-retrieval baselines include BM25, Cosine Similarity with \texttt{text-embedding-3-small}, and TopicK~\citep{kweon2025topick}. Dynamic-retrieval baselines include Memento~\citep{zhou2025memento} and MemRL~\citep{zhang2026memrl}, which learn memory selection or utility but operate primarily at the episode level. For applicable memory-based methods, we keep the episode-level memory writing mechanism identical to isolate the effect of retrieval.

\paragraph{Model Configurations.}
The base reasoning agent uses GPT-5~\citep{singh2025openai} with low reasoning effort, and the user simulator uses GPT-4o~\citep{hurst2024gpt}. The PRM is initialized from \textit{DeepSeek-R1-Distill-Qwen-7B}~\citep{guo2025deepseek}, and the retrieval selectors use LoRA-tuned \textit{Qwen3-4B-Instruct-2507}~\citep{yang2025qwen3}. We use 5,346 teacher-annotated state-memory pairs for PRM SFT, 42,768 pairs for GRPO, and 5,346 pairs for PRM evaluation. 

More details are provided in Appendix~\ref{appendix: experimental settings}.


\subsection{Experimental Results}

\paragraph{Main Results.}
Table~\ref{tab: main results} compares \ours{} with no-memory, static-retrieval, and dynamic-retrieval baselines on BIRD-Interact.
\ours{} achieves the best overall success rates in both Phase~1 and Phase~2, reaching 33.01\% and 22.17\%, respectively. 
The gains hold across both READ and WRITE tasks, indicating that the learned retrieval policies help both analytical extraction and state-changing operations.
Furthermore, \ours{} achieves the lowest average number of interaction turns. These results indicate that \ours{} improves task success while reducing interaction turns.
To verify that this ranking is stable, we repeat \ours{} and the strongest dynamic baseline five times in Appendix~\ref{app:variance_analysis}, where \ours{}'s weakest run still exceeds MemRL's best run in both phases.

\paragraph{Transferability.}
Table~\ref{tab: spider results} evaluates whether retrieval policies trained on BIRD-Interact transfer to Spider2-Snow without Spider2-Snow-specific training or tuning. \ours{} achieves the highest success rate among all methods, outperforming the strongest baseline. The gain is smaller than on BIRD-Interact, as expected under cross-benchmark transfer. 
Nevertheless, this suggests that \ours{}'s learned policies transfer without degradation to a new benchmark.
\ours{} also keeps average turns close to the No Memory Agent and below all memory-retrieval baselines, indicating improved success without substantial interaction cost.







\begin{figure*}[t]
    \centering

    \begin{subfigure}[t]{0.495\linewidth}
        \centering
        \includegraphics[width=\linewidth]{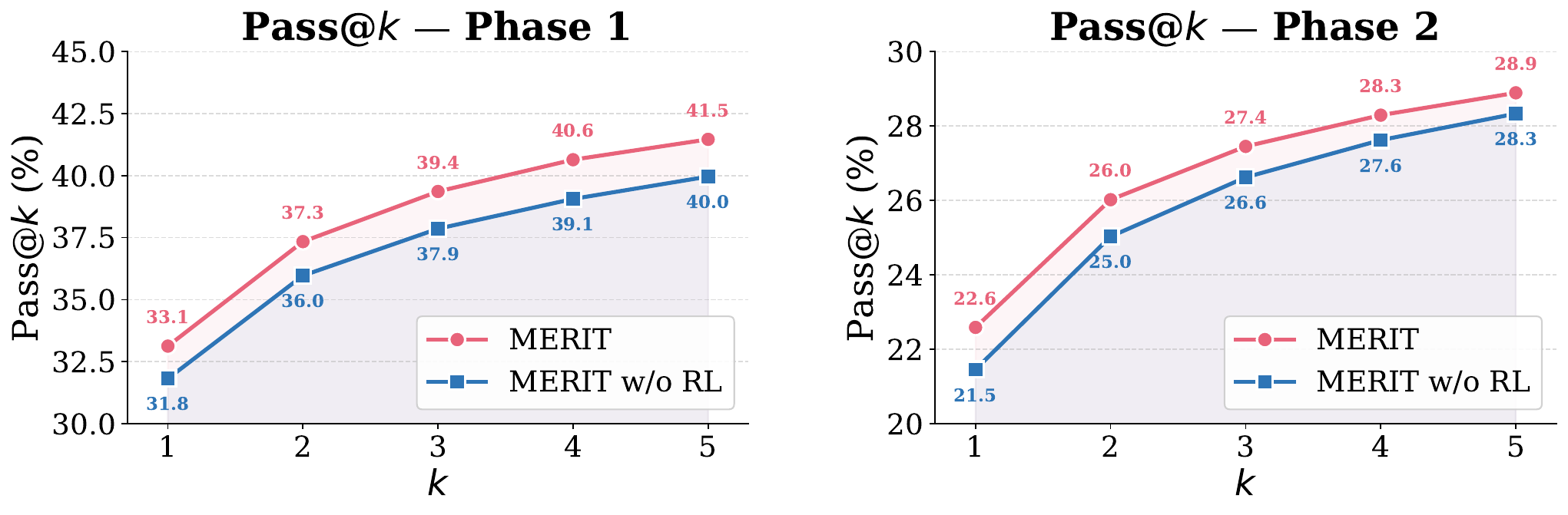}
        \caption{Pass@$k$ analysis.}
        \label{fig:passk}
    \end{subfigure}
    \hfill
    \begin{subfigure}[t]{0.495\linewidth}
        \centering
        \includegraphics[width=\linewidth]{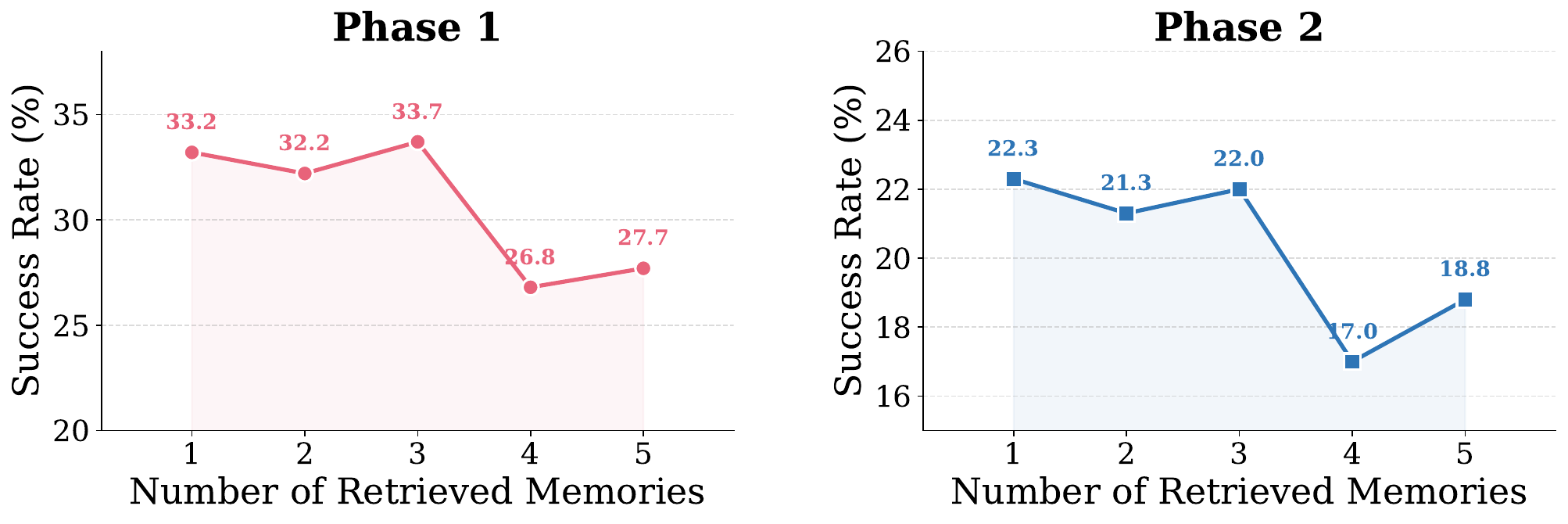}
        \caption{Number of retrieved memories.}
        \label{fig:number_memories}
    \end{subfigure}

    \caption{Retrieval robustness and retrieval size. 
    MERIT consistently outperforms the non-RL variant across $k$, while the retrieval-size analysis shows that injecting too many memories can introduce noisy or redundant guidance.}
    \label{fig:retrieval_ablation}
\end{figure*}
\begin{table*}[t]
\centering
\resizebox{\linewidth}{!}{%
\begin{tabular}{lccccccccc}
\toprule
\multirow{2}{*}{\textbf{Methods}} & \multicolumn{3}{c}{\textbf{Phase 1}} & \multicolumn{3}{c}{\textbf{Phase 2}} & \multicolumn{3}{c}{\textbf{Avg. \#Turns}$\downarrow$} \\
\cmidrule(lr){2-4} \cmidrule(lr){5-7} \cmidrule(lr){8-10}
& \textbf{WRITE} & \textbf{READ} & \textbf{Overall} & \textbf{WRITE} & \textbf{READ} & \textbf{Overall} & \textbf{WRITE} & \textbf{READ} & \textbf{Overall} \\
\midrule
\ours{} (ours)  & \textbf{53.49}&\textbf{24.64}&\textbf{33.01}&\textbf{34.30}&\textbf{17.22}&\textbf{22.17}&\underline{7.09}&\underline{8.15}&\underline{7.84} \\
\ w/o episode & 42.53&15.73&23.50&21.84&8.92&12.70&8.10&8.88&8.65\\
\ w/o turn & 48.85&20.19&28.50&28.74&12.44&17.20&7.67&8.68&8.39\\
\ w/o RL & \underline{52.87}&\underline{24.18}&\underline{32.50}&\underline{29.89}&\underline{15.26}&\underline{19.50}&\textbf{7.07}&\textbf{8.07}&\textbf{7.78}\\
\bottomrule
\end{tabular}
}
\caption{Ablation studies evaluating the contribution of individual framework components, including episode-level strategy, turn-level hints, and RL optimization, on task success rates and interaction efficiency.}
\label{tab: ablation}
\end{table*}

\paragraph{Ablation Studies.}
Table~\ref{tab: ablation} ablates each \ours{} component. Removing episode-level memory causes the largest drop, reducing overall success to 23.50\% in Phase~1 and 12.70\% in Phase~2. This shows that episode-level experience is important in interactive text-to-SQL. Removing turn-level memory also degrades performance, reducing overall success to 28.50\% and 17.20\% in the two phases and increasing the average number of turns to 8.39. This suggests that turn-level hints complement episode-level memories by supporting decisions as the interaction state evolves.
Finally, removing RL optimization has a modest effect on Phase 1 but lowers Phase 2 success from 22.17\% to 19.50\%. This suggests that utility-based selector optimization is most beneficial for state-dependent follow-up interactions, where memory usefulness depends more strongly on the evolving context.


\begin{table}[t]
\centering
\resizebox{\linewidth}{!}{%
\begin{tabular}{lcc}
\toprule
\textbf{Condition} & \textbf{\% Next Action Valid} & \textbf{\% Final Success} \\
\midrule
\ours{} (ours) & 60.0 & \textbf{20.0} \\
BM25 & \textbf{62.5} & 10.0 \\
No Memory & 55.0 & 7.5 \\
\bottomrule
\end{tabular}
}
\caption{Counterfactual analysis at a frozen intermediate state ($k=7$)
over 40 episodes. BM25 attains a higher next-action-valid rate, yet the
\ours{} selector yields a higher final success rate, showing that locally
plausible memories are not necessarily globally correct.}
\label{tab: counterfactual}
\end{table}
\paragraph{Counterfactual Analysis.}
To isolate the effect of a single turn-level memory, we run a counterfactual at a fixed state. We sample 40 episodes, run each to turn $k=7$, freeze the state, and fork three continuations that differ only in the injected turn-level memory, from the \ours{} selector, a BM25 baseline, and no memory. We report next-action validity and final success in Table~\ref{tab: counterfactual}. \ours{} attains a slightly lower next-action-valid rate than BM25 but a higher final success rate, and both exceed no memory. BM25 favors surface-similar memories that make the immediate action executable but not the final query correct, whereas the learned selector favors downstream usefulness and converts more episodes into correct submissions. This suggests the \ours{} turn-level selector chooses memories with positive utility toward the outcome rather than mere local relevance.

\begin{table}[t]
\begin{center}
\resizebox{\linewidth}{!}{
\begin{tabular}{lcccc}
\toprule
\textbf{Training Stage} & \textbf{Valid (\%)} & \textbf{avg. Acc} & \textbf{avg. MAE} & \textbf{avg. MSE}\\
\midrule
Vanilla & 95.59 & 33.67 & 0.94 & 1.73\\
SFT & 92.12 & 36.84 & 0.85 & 1.50 \\
SFT + GRPO & 100.00 & 73.59 & 0.29 & 0.34 \\
\bottomrule
\end{tabular}
}
\end{center}
\caption{Performance of PRM.}
\label{tab: prm_results}
\end{table}

\paragraph{PRM Evaluation.}
Table~\ref{tab: prm_results} evaluates whether the PRM can produce structured utility judgments aligned with teacher annotations. Compared with the vanilla base model and the SFT-only model, SFT+GRPO achieves 100\% valid formatting, higher annotation agreement, and lower numeric-score errors. Specifically, SFT+GRPO improves average accuracy to 73.59\% and reduces MAE and MSE to 0.29 and 0.34, respectively.
These results show that GRPO improves agreement with teacher utility rubrics; downstream results test whether such rewards help retrieval.
Since direct intermediate utility labels are unavailable, we use this teacher-aligned score as a structured proxy reward and examine its downstream effects through turn-level and RL ablations.

\paragraph{Retrieval Behavior Analysis.}
We further analyze the robustness of learned retrieval and the effect of retrieval size.
Figure~\ref{fig:passk} compares \ours{} with its non-RL variant under different Pass@\(k\) budgets. \ours{} achieves higher success across all \(k\), indicating that RL-based retrieval remains beneficial across different generation budgets. 
Figure~\ref{fig:number_memories} studies the number of retrieved memories. 
Retrieving one to three memories yields comparable performance, whereas larger retrieval budgets reduce success rates. This suggests that excessive memory context may introduce redundant or distracting information.
We provide an additional analysis of memory composition in Appendix~\ref{app:memory_composition}.

\begin{table}[t]
\centering
\small
\begin{tabular}{lcc}
\toprule
\textbf{Methods} & \textbf{Avg. \#Turns} & \textbf{\#Tokens} \\
\midrule
\textbf{MERIT (ours)} & 7.84 & 100.4M  \\
\ w/o episode & 8.65 & 95.4M  \\
\ w/o turn & 8.39 & 108.5M  \\
\ w/o RL & 7.78 & 104.2M  \\
\bottomrule
\end{tabular}
\caption{Token and interaction efficiency comparison.}
\label{tab:token_efficiency_ablation}
\end{table}
\paragraph{Computational and Token Efficiency.}
Although \ours{} introduces learned retrieval modules, the additional cost is limited because the selectors are lightweight and operate on compact state signatures and memory summaries. Table~\ref{tab:token_efficiency_ablation} compares token usage and interaction turns across ablations. \ours{} uses 100.4M tokens, fewer than the variants without turn-level memory or RL optimization, which require 108.5M and 104.2M tokens, respectively. The variant without episode-level memory uses fewer tokens, yet it substantially reduces success rates and increases interaction turns. Overall, \ours{} offers a favorable success-efficiency trade-off by improving task performance while reducing unnecessary interactions and tokens.

\paragraph{Leaderboard Results. }
We additionally transfer the retrieval artifacts trained with GPT-5 as the reasoning agent, GPT-4o as the user simulator, and the same Qwen-based retrievers used in the main experiments to two stronger agent configurations. Using GPT-5.4 as the reasoning agent with GPT-4o as the user simulator, \ours{} ranks first on both Phase 1 and Phase 2 of the BIRD-Interact leaderboard\footnote{\url{https://bird-interact.github.io/}} as of August 2026 (see Appendix~\ref{app:leaderboard}). A second configuration with Claude-Opus-4.6 as the reasoning agent and Claude-Haiku-4-5 as the user simulator ranks second on Phase 2, where adding \ours{}'s retrieval raises success from 17.50 to 20.83 over the base Claude-Opus-4.6 agent. 












\section{Conclusion}
We proposed \ours{}, a dual-level long-term memory framework that learns episode-level and turn-level retrieval policies for interactive text-to-SQL agents. By separating global strategic guidance from local state-conditioned hints and using a lightweight PRM to provide dense rewards for turn-level retrieval, \ours{} enables horizon-aware memory selection. Experiments on BIRD-Interact show improved success rates with fewer interaction turns, while Spider2-Snow results provide evidence of transfer beyond the source benchmark.

\section*{Limitations}

First, the PRM is trained from teacher-generated rubrics rather than ground-truth labels of turn-level memory utility. This is a practical limitation of interactive retrieval settings, where the usefulness of a memory at an intermediate turn is difficult to observe directly because the agent state, future actions, and environment feedback all evolve after the retrieval decision. We therefore use a strong teacher model to provide structured utility annotations as a proxy supervision signal. While this does not provide causal ground-truth attribution for each retrieved memory, the downstream results and ablation studies suggest that the PRM-trained turn-level selector improves state-conditioned retrieval and contributes to better follow-up interaction performance.

Second, our current evaluation uses a strong proprietary backbone for the main agent and teacher annotation. This setting allows us to study memory retrieval under a capable interactive agent, but it leaves open whether the same gains hold across open-source agent backbones or weaker reasoning models. Evaluating \ours{} with different base agents, teacher models, and simulator configurations is an important direction for future work.


\bibliography{colm2026_conference}

\begin{thebibliography}{39}
\providecommand{\natexlab}[1]{#1}

\bibitem[{Arslan et~al.(2024)Arslan, Ghanem, Munawar, and Cruz}]{arslan2024survey}
Muhammad Arslan, Hussam Ghanem, Saba Munawar, and Christophe Cruz. 2024.
\newblock \href {https://doi.org/10.1016/j.procs.2024.09.178} {A survey on rag with llms}.
\newblock \emph{Procedia Computer Science}, 246:3781--3790.
\newblock 28th International Conference on Knowledge Based and Intelligent information and Engineering Systems (KES 2024).

\bibitem[{Asai et~al.(2024)Asai, Wu, Wang, Sil, and Hajishirzi}]{ICLR2024_25f7be96}
Akari Asai, Zeqiu Wu, Yizhong Wang, Avi Sil, and Hannaneh Hajishirzi. 2024.
\newblock \href {https://proceedings.iclr.cc/paper_files/paper/2024/file/25f7be9694d7b32d5cc670927b8091e1-Paper-Conference.pdf} {Self-rag: Learning to retrieve, generate, and critique through self-reflection}.
\newblock In \emph{International Conference on Learning Representations}, volume 2024, pages 9112--9141.

\bibitem[{Chhikara et~al.(2025)Chhikara, Khant, Aryan, Singh, and Yadav}]{chhikara2025mem0}
Prateek Chhikara, Dev Khant, Saket Aryan, Taranjeet Singh, and Deshraj Yadav. 2025.
\newblock \href {https://doi.org/10.3233/faia251160} {Mem0: Building production-ready ai agents with scalable long-term memory}.
\newblock In \emph{ECAI 2025}. IOS Press.

\bibitem[{Choudhury(2025)}]{choudhury2025process}
Sanjiban Choudhury. 2025.
\newblock \href {https://arxiv.org/abs/2502.10325} {Process reward models for llm agents: Practical framework and directions}.
\newblock \emph{Preprint}, arXiv:2502.10325.

\bibitem[{Fang et~al.(2026)Fang, Liang, Wang, Wu, Qiao, Xie, Huang, Chen, and Zhang}]{fang2025memp}
Runnan Fang, Yuan Liang, Xiaobin Wang, Jialong Wu, Shuofei Qiao, Pengjun Xie, Fei Huang, Huajun Chen, and Ningyu Zhang. 2026.
\newblock \href {https://doi.org/10.18653/v1/2026.findings-acl.866} {Memp: Exploring agent procedural memory}.
\newblock In \emph{Findings of the {A}ssociation for {C}omputational {L}inguistics: {ACL} 2026}, pages 17490--17502, San Diego, California, United States. Association for Computational Linguistics.

\bibitem[{Guo et~al.(2025)Guo, Yang, Zhang, Song, Wang, Zhu, Xu, Zhang, Ma, Bi, Zhang, Yu, Wu, Wu, Gou, Shao, Li, Gao, Liu, Xue, Wang, Wu, Feng, Lu, Zhao, Deng, Ruan, Dai, Chen, Ji, Li, Lin, Dai, Luo, Hao, Chen, Li, Zhang, Xu, Ding, Gao, Qu, Li, Guo, Li, Chen, Yuan, Tu, Qiu, Li, Cai, Ni, Liang, Chen, Dong, Hu, You, Gao, Guan, Huang, Yu, Wang, Zhang, Zhao, Wang, Zhang, Xu, Xia, Zhang, Zhang, Tang, Zhou, Li, Wang, Li, Tian, Huang, Zhang, Wang, Chen, Du, Ge, Zhang, Pan, Wang, Chen, Jin, Chen, Lu, Zhou, Chen, Ye, Wang, Yu, Zhou, Pan, Li, Zhou, Wu, Yun, Pei, Sun, Wang, Zeng, Liu, Liang, Gao, Yu, Zhang, Xiao, An, Liu, Wang, Chen, Nie, Cheng, Liu, Xie, Liu, Yang, Li, Su, Lin, Li, Jin, Shen, Chen, Sun, Wang, Song, Zhou, Wang, Shan, Li, Wang, Wei, Zhang, Xu, Li, Zhao, Sun, Wang, Yu, Zhang, Shi, Xiong, He, Piao, Wang, Tan, Ma, Liu, Guo, Ou, Wang, Gong, Zou, He, Xiong, Luo, You, Liu, Zhou, Zhu, Huang, Li, Zheng, Zhu, Ma, Tang, Zha, Yan, Ren, Ren, Sha, Fu, Xu, Xie, Zhang, Hao, Ma, Yan, Wu, Gu, Zhu, Liu, Li, Xie, Song,
  Pan, Huang, Xu, Zhang, and Zhang}]{guo2025deepseek}
Daya Guo, Dejian Yang, Haowei Zhang, Junxiao Song, Peiyi Wang, Qihao Zhu, Runxin Xu, Ruoyu Zhang, Shirong Ma, Xiao Bi, Xiaokang Zhang, Xingkai Yu, Yu~Wu, Z.~F. Wu, Zhibin Gou, Zhihong Shao, Zhuoshu Li, Ziyi Gao, Aixin Liu, and 175 others. 2025.
\newblock \href {https://doi.org/10.1038/s41586-025-09422-z} {Deepseek-r1 incentivizes reasoning in llms through reinforcement learning}.
\newblock \emph{Nature}, 645(8081):633–638.

\bibitem[{Huo et~al.(2026)Huo, Xu, Li, Jacobsson, Lin, Qin, Hui, Li, Qu, Si, Han, Alexander, Zhu, Qin, Yu, Jin, Zhou, Zhong, Chen, Liu, Ma, Ozcan, Papakonstantinou, and Cheng}]{huo2025bird}
Nan Huo, Xiaohan Xu, Jinyang Li, Per Jacobsson, Shipei Lin, Bowen Qin, Binyuan Hui, Xiaolong Li, Ge~Qu, Shuzheng Si, Linheng Han, Edward Alexander, Xintong Zhu, Rui Qin, Ruihan Yu, Yiyao Jin, Feige Zhou, Weihao Zhong, Yun Chen, and 5 others. 2026.
\newblock \href {https://openreview.net/forum?id=nHrYBGujps} {{BIRD}-{INTERACT}: Re-imagining text-to-{SQL} evaluation via lens of dynamic interactions}.
\newblock In \emph{The Fourteenth International Conference on Learning Representations}.

\bibitem[{Hurst et~al.(2024)Hurst, Lerer, Goucher, Perelman, Ramesh, Clark, Ostrow, Welihinda, Hayes, Radford, Mądry, Baker-Whitcomb, Beutel, Borzunov, Carney, Chow, Kirillov, Nichol, Paino, Renzin, Passos, Kirillov, Christakis, Conneau, Kamali, Jabri, Moyer, Tam, Crookes, Tootoochian, Tootoonchian, Kumar, Vallone, Karpathy, Braunstein, Cann, Codispoti, Galu, Kondrich, Tulloch, Mishchenko, Baek, Jiang, Pelisse, Woodford, Gosalia, Dhar, Pantuliano, Nayak, Oliver, Zoph, Ghorbani, Leimberger, Rossen, Sokolowsky, Wang, Zweig, Hoover, Samic, McGrew, Spero, Giertler, Cheng, Lightcap, Walkin, Quinn, Guarraci, Hsu, Kellogg, Eastman, Lugaresi, Wainwright, Bassin, Hudson, Chu, Nelson, Li, Shern, Conger, Barette, Voss, Ding, Lu, Zhang, Beaumont, Hallacy, Koch, Gibson, Kim, Choi, McLeavey, Hesse, Fischer, Winter, Czarnecki, Jarvis, Wei, Koumouzelis, Sherburn, Kappler, Levin, Levy, Carr, Farhi, Mely, Robinson, Sasaki, Jin, Valladares, Tsipras, Li, Nguyen, Findlay, Oiwoh, Wong, Asdar, Proehl, Yang, Antonow, Kramer,
  Peterson, Sigler, Wallace, Brevdo, Mays, Khorasani, Such, Raso, Zhang, von Lohmann, Sulit, Goh, Oden, Salmon, Starace, Brockman, Salman, Bao, Hu, Wong, Wang, Schmidt, Whitney, Jun, Kirchner, de~Oliveira~Pinto, Ren, Chang, Chung, Kivlichan, O'Connell, O'Connell, Osband, Silber, Sohl, Okuyucu, Lan, Kostrikov, Sutskever, Kanitscheider, Gulrajani, Coxon, Menick, Pachocki, Aung, Betker, Crooks, Lennon, Kiros, Leike, Park, Kwon, Phang, Teplitz, Wei, Wolfe, Chen, Harris, Varavva, Lee, Shieh, Lin, Yu, Weng, Tang, Yu, Jang, Candela, Beutler, Landers, Parish, Heidecke, Schulman, Lachman, McKay, Uesato, Ward, Kim, Huizinga, Sitkin, Kraaijeveld, Gross, Kaplan, Snyder, Achiam, Jiao, Lee, Zhuang, Harriman, Fricke, Hayashi, Singhal, Shi, Karthik, Wood, Rimbach, Hsu, Nguyen, Gu-Lemberg, Button, Liu, Howe, Muthukumar, Luther, Ahmad, Kai, Itow, Workman, Pathak, Chen, Jing, Guy, Fedus, Zhou, Mamitsuka, Weng, McCallum, Held, Ouyang, Feuvrier, Zhang, Kondraciuk, Kaiser, Hewitt, Metz, Doshi, Aflak, Simens, Boyd, Thompson,
  Dukhan, Chen, Gray, Hudnall, Zhang, Aljubeh, Litwin, Zeng, Johnson, Shetty, Gupta, Shah, Yatbaz, Yang, Zhong, Glaese, Chen, Janner, Lampe, Petrov, Wu, Wang, Fradin, Pokrass, Castro, de~Castro, Pavlov, Brundage, Wang, Khan, Murati, Bavarian, Lin, Yesildal, Soto, Gimelshein, Cone, Staudacher, Summers, LaFontaine, Chowdhury, Ryder, Stathas, Turley, Tezak, Felix, Kudige, Keskar, Deutsch, Bundick, Puckett, Nachum, Okelola, Boiko, Murk, Jaffe, Watkins, Godement, Campbell-Moore, Chao, McMillan, Belov, Su, Bak, Bakkum, Deng, Dolan, Hoeschele, Welinder, Tillet, Pronin, Tillet, Dhariwal, Yuan, Dias, Lim, Arora, Troll, Lin, Lopes, Puri, Miyara, Leike, Gaubert, Zamani, Wang, Donnelly, Honsby, Smith, Sahai, Ramchandani, Huet, Carmichael, Zellers, Chen, Chen, Nigmatullin, Cheu, Jain, Altman, Schoenholz, Toizer, Miserendino, Agarwal, Culver, Ethersmith, Gray, Grove, Metzger, Hermani, Jain, Zhao, Wu, Jomoto, Wu, Shuaiqi, Xia, Phene, Papay, Narayanan, Coffey, Lee, Hall, Balaji, Broda, Stramer, Xu, Gogineni, Christianson,
  Sanders, Patwardhan, Cunninghman, Degry, Dimson, Raoux, Shadwell, Zheng, Underwood, Markov, Sherbakov, Rubin, Stasi, Kaftan, Heywood, Peterson, Walters, Eloundou, Qi, Moeller, Monaco, Kuo, Fomenko, Chang, Zheng, Zhou, Manassra, Sheu, Zaremba, Patil, Qian, Kim, Cheng, Zhang, He, Zhang, Jin, Dai, and Malkov}]{hurst2024gpt}
Aaron Hurst, Adam Lerer, Adam~P. Goucher, Adam Perelman, Aditya Ramesh, Aidan Clark, AJ~Ostrow, Akila Welihinda, Alan Hayes, Alec Radford, Aleksander Mądry, Alex Baker-Whitcomb, Alex Beutel, Alex Borzunov, Alex Carney, Alex Chow, Alex Kirillov, Alex Nichol, Alex Paino, and 399 others. 2024.
\newblock \href {https://arxiv.org/abs/2410.21276} {Gpt-4o system card}.
\newblock \emph{Preprint}, arXiv:2410.21276.

\bibitem[{Kagaya et~al.(2024)Kagaya, Yuan, Lou, Karlekar, Pranata, Kinose, Oguri, Wick, and You}]{kagaya2024rap}
Tomoyuki Kagaya, Thong~Jing Yuan, Yuxuan Lou, Jayashree Karlekar, Sugiri Pranata, Akira Kinose, Koki Oguri, Felix Wick, and Yang You. 2024.
\newblock \href {https://arxiv.org/abs/2402.03610} {Rap: Retrieval-augmented planning with contextual memory for multimodal llm agents}.
\newblock \emph{Preprint}, arXiv:2402.03610.

\bibitem[{Kweon et~al.(2025)Kweon, Kang, Tian, Jiang, Han, and Yu}]{kweon2025topick}
Wonbin Kweon, SeongKu Kang, Runchu Tian, Pengcheng Jiang, Jiawei Han, and Hwanjo Yu. 2025.
\newblock \href {https://doi.org/10.18653/v1/2025.emnlp-main.1007} {Topic coverage-based demonstration retrieval for in-context learning}.
\newblock In \emph{Proceedings of the 2025 Conference on Empirical Methods in Natural Language Processing}, pages 19900--19912, Suzhou, China. Association for Computational Linguistics.

\bibitem[{Lei et~al.(2025)Lei, Chen, Ye, Cao, Shin, SU, SUO, Gao, Hu, Yin, Zhong, Xiong, Sun, Liu, Wang, and Yu}]{lei2025spider}
Fangyu Lei, Jixuan Chen, Yuxiao Ye, Ruisheng Cao, Dongchan Shin, Hongjin SU, ZHAOQING SUO, Hongcheng Gao, Wenjing Hu, Pengcheng Yin, Victor Zhong, Caiming Xiong, Ruoxi Sun, Qian Liu, Sida Wang, and Tao Yu. 2025.
\newblock \href {https://openreview.net/forum?id=XmProj9cPs} {Spider 2.0: Evaluating language models on real-world enterprise text-to-{SQL} workflows}.
\newblock In \emph{The Thirteenth International Conference on Learning Representations}.

\bibitem[{Li(2025)}]{li2025review}
Xinzhe Li. 2025.
\newblock \href {https://aclanthology.org/2025.coling-main.652/} {A review of prominent paradigms for {LLM}-based agents: Tool use, planning (including {RAG}), and feedback learning}.
\newblock In \emph{Proceedings of the 31st International Conference on Computational Linguistics}, pages 9760--9779, Abu Dhabi, UAE. Association for Computational Linguistics.

\bibitem[{Lin et~al.(2020)Lin, Socher, and Xiong}]{lin-etal-2020-bridging}
Xi~Victoria Lin, Richard Socher, and Caiming Xiong. 2020.
\newblock \href {https://doi.org/10.18653/v1/2020.findings-emnlp.438} {Bridging textual and tabular data for cross-domain text-to-{SQL} semantic parsing}.
\newblock In \emph{Findings of the Association for Computational Linguistics: EMNLP 2020}, pages 4870--4888, Online. Association for Computational Linguistics.

\bibitem[{Liu et~al.(2026)Liu, Su, Xia, Han, Zheng, Xie, Ding, and Yao}]{liu2026simplemem}
Jiaqi Liu, Yaofeng Su, Peng Xia, Siwei Han, Zeyu Zheng, Cihang Xie, Mingyu Ding, and Huaxiu Yao. 2026.
\newblock \href {https://openreview.net/forum?id=oBgLvd5YC6} {Simplemem: Efficient lifelong memory for {LLM} agents}.
\newblock In \emph{Forty-third International Conference on Machine Learning}.

\bibitem[{Lu et~al.(2023)Lu, An, Lin, Pergola, He, Yin, Sun, and Wu}]{lu2023memochat}
Junru Lu, Siyu An, Mingbao Lin, Gabriele Pergola, Yulan He, Di~Yin, Xing Sun, and Yunsheng Wu. 2023.
\newblock \href {https://arxiv.org/abs/2308.08239} {Memochat: Tuning llms to use memos for consistent long-range open-domain conversation}.
\newblock \emph{Preprint}, arXiv:2308.08239.

\bibitem[{Maharana et~al.(2024)Maharana, Lee, Tulyakov, Bansal, Barbieri, and Fang}]{maharana2024evaluating}
Adyasha Maharana, Dong-Ho Lee, Sergey Tulyakov, Mohit Bansal, Francesco Barbieri, and Yuwei Fang. 2024.
\newblock \href {https://doi.org/10.18653/v1/2024.acl-long.747} {Evaluating very long-term conversational memory of {LLM} agents}.
\newblock In \emph{Proceedings of the 62nd Annual Meeting of the Association for Computational Linguistics (Volume 1: Long Papers)}, pages 13851--13870, Bangkok, Thailand. Association for Computational Linguistics.

\bibitem[{Mohammadi et~al.(2025)Mohammadi, Li, Lo, and Yip}]{mohammadi2025evaluation}
Mahmoud Mohammadi, Yipeng Li, Jane Lo, and Wendy Yip. 2025.
\newblock \href {https://doi.org/10.1145/3711896.3736570} {Evaluation and benchmarking of llm agents: A survey}.
\newblock In \emph{Proceedings of the 31st ACM SIGKDD Conference on Knowledge Discovery and Data Mining V.2}, KDD '25, page 6129–6139, New York, NY, USA. Association for Computing Machinery.

\bibitem[{Packer et~al.(2024)Packer, Wooders, Lin, Fang, Patil, Stoica, and Gonzalez}]{packer2023memgpt}
Charles Packer, Sarah Wooders, Kevin Lin, Vivian Fang, Shishir~G. Patil, Ion Stoica, and Joseph~E. Gonzalez. 2024.
\newblock \href {https://arxiv.org/abs/2310.08560} {Memgpt: Towards llms as operating systems}.
\newblock \emph{Preprint}, arXiv:2310.08560.

\bibitem[{Park et~al.(2023)Park, O'Brien, Cai, Morris, Liang, and Bernstein}]{park2023generative}
Joon~Sung Park, Joseph O'Brien, Carrie~Jun Cai, Meredith~Ringel Morris, Percy Liang, and Michael~S. Bernstein. 2023.
\newblock \href {https://doi.org/10.1145/3586183.3606763} {Generative agents: Interactive simulacra of human behavior}.
\newblock In \emph{Proceedings of the 36th Annual ACM Symposium on User Interface Software and Technology}, UIST '23, New York, NY, USA. Association for Computing Machinery.

\bibitem[{Salama et~al.(2025)Salama, Cai, Yuan, Currey, Sunkara, Zhang, and Benajiba}]{salama2025meminsight}
Rana Salama, Jason Cai, Michelle Yuan, Anna Currey, Monica Sunkara, Yi~Zhang, and Yassine Benajiba. 2025.
\newblock \href {https://doi.org/10.18653/v1/2025.emnlp-main.1683} {{M}em{I}nsight: Autonomous memory augmentation for {LLM} agents}.
\newblock In \emph{Proceedings of the 2025 Conference on Empirical Methods in Natural Language Processing}, pages 33136--33152, Suzhou, China. Association for Computational Linguistics.

\bibitem[{Shao et~al.(2025)Shao, Cai, Lin, and Ming}]{shao2025enhancing}
Zhihui Shao, Shubin Cai, Rongsheng Lin, and Zhong Ming. 2025.
\newblock \href {https://doi.org/10.18653/v1/2025.findings-naacl.245} {Enhancing text-to-{SQL} with question classification and multi-agent collaboration}.
\newblock In \emph{Findings of the Association for Computational Linguistics: NAACL 2025}, pages 4340--4349, Albuquerque, New Mexico. Association for Computational Linguistics.

\bibitem[{Singh et~al.(2026)Singh, Fry, Perelman, Tart, Ganesh, El-Kishky, McLaughlin, Low, Ostrow, Ananthram, Nathan, Luo, Helyar, Madry, Efremov, Spyra, Baker-Whitcomb, Beutel, Karpenko, Makelov, Neitz, Wei, Barr, Kirchmeyer, Ivanov, Christakis, Gillespie, Tam, Bennett, Wan, Huang, Sandjideh, Yang, Kumar, Saraiva, Vallone, Gheorghe, Garcia, Braunstein, Liu, Schmidt, Mereskin, Mishchenko, Applebaum, Rogerson, Rajan, Wei, Kotha, Srivastava, Agrawal, Vijayvergiya, Tyra, Nair, Nayak, Eggers, Ji, Hoover, Chen, Chen, Barak, Minaiev, Hao, Baker, Lightcap, McKinzie, Wang, Quinn, Fioca, Hsu, Yang, Yu, Zhang, Brenner, Zetino, Raymond, Lugaresi, Paz, Hudson, Whitney, Li, Chen, Cole, Voss, Ding, Shen, Huang, Colby, Hallacy, Koch, Lu, Kaplan, Kim, Minott-Henriques, Frey, Yu, Czarnecki, Reid, Wei, Decareaux, Scheau, Zhang, Forbes, Tang, Goldberg, Roberts, Palmie, Kappler, Levine, Wright, Leo, Lin, Robinson, Grabb, Chen, Lim, Salama, Bhattacharjee, Tsipras, Li, Yu, Strouse, Williams, Hunn, Bayes, Arbus, Akyurek, Le,
  Widmann, Yani, Proehl, Sert, Cheung, Schwartz, Han, Jiang, Mitchell, Sigler, Wallace, Ritter, Kavanaugh, Mays, Nikishin, Li, Such, de~Avila Belbute~Peres, Raso, Bekerman, Tsimpourlas, Chantzis, Song, Zhang, Raila, McGrath, Briggs, Yang, Parascandolo, Chabot, Kim, Zhao, Valiant, Leclerc, Salman, Wang, Sheng, Jiang, Wang, Jin, Sikchi, Schmidt, Aspegren, Chen, Qiu, Lightman, Covert, Kivlichan, Silber, Sohl, Hammoud, Clavera, Lan, Akkaya, Kostrikov, Kofman, Etinger, Singal, Hehir, Huh, Pan, Wilczynski, Pachocki, Lee, Quinn, Kiros, Kalra, Samaroo, Wang, Wolfe, Chen, Wang, Harb, Han, Wang, Zhao, Chen, Yang, Tworek, Chand, Landon, Liang, Lin, Liu, Wang, Tang, Yin, Jang, Morris, Flynn, Ferstad, Heidecke, Fishbein, Hallman, Grant, Chien, Gordon, Park, Liss, Kraaijeveld, Guay, Mo, Lawson, McGrath, Vendrow, Jiao, Lee, Steele, Wang, Mao, Chen, Hayashi, Xiao, Salahi, Wu, Sekhri, Sharma, Singhal, Li, Nguyen, Gu-Lemberg, King, Liu, Stone, Yu, Ying, Georgiev, Lim, Tirumala, Miller, Ahmad, Lv, Clare, Fauconnet, Itow, Yang,
  Romaniuk, Anise, Byron, Pathak, Maksin, Lo, Ho, Jing, Wu, Xiong, Mamitsuka, Yang, McCallum, Held, Bourgeois, Engstrom, Kuhn, Feuvrier, Zhang, Switzer, Kondraciuk, Kaiser, Joglekar, Singh, Shah, Stratta, Williams, Chen, Sun, Cayton, Li, Zhang, Aljubeh, Nichols, Haines, Schwarzer, Gupta, Shah, Guan, Huang, Dong, Wang, Glaese, Carroll, Lampe, Malek, Sharman, Zhang, Wang, Pokrass, Florian, Pavlov, Wang, Chen, Wang, Feng, Bavarian, Lin, Abdool, Rohaninejad, Soto, Staudacher, LaFontaine, Marwell, Liu, Preston, Turley, Ansman, Blades, Pancha, Mikhaylin, Felix, Handa, Rai, Keskar, Brown, Nachum, Boiko, Murk, Watkins, Gleeson, Mishkin, Lesiewicz, Baltescu, Belov, Zhokhov, Pronin, Guo, Thacker, Liu, Yuan, Liu, Dias, Puckett, Arora, Mullapudi, Gaon, Miyara, Song, Aggarwal, Marsan, Yemiru, Xiong, Kshirsagar, Nuttall, Tsiupa, Eldan, Wang, James, Ziv, Shu, Nigmatullin, Jain, Talaie, Altman, Arnesen, Toizer, Toyer, Miserendino, Agarwal, Yoo, Heon, Ethersmith, Grove, Taylor, Bubeck, Banesiu, Amdo, Zhao, Wu, Santurkar,
  Zhao, Chaudhuri, Krishnaswamy, Shuaiqi, Xia, Cheng, Anadkat, Fishman, Tobin, Fu, Jain, Mei, Egoian, Kim, Golden, Mah, Lin, Imm, Sharpe, Yadlowsky, Choudhry, Eum, Sanjeev, Khan, Stramer, Wang, Xin, Gogineni, Christianson, Sanders, Patwardhan, Degry, Shadwell, Fu, Gao, Garipov, Sriskandarajah, Sherbakov, Korbak, Kaftan, Hiratsuka, Wang, Song, Zhao, Peterson, Kharitonov, Chernova, Kosaraju, Kuo, Pong, Verma, Petrov, Jiang, Zhang, Zhou, Xie, Zhan, McCabe, DePue, Ellsworth, Bain, Thompson, Chen, Qi, Xiang, Shi, Dubois, Yu, Khakbaz, Wu, Qian, Lee, Chen, Zhang, Xiong, Tian, Cha, Bai, Yang, Yuan, Li, Zhang, Yang, Jin, Jiang, Wang, Wang, Liu, Stubenvoll, Dou, Wu, and Wang}]{singh2025openai}
Aaditya Singh, Adam Fry, Adam Perelman, Adam Tart, Adi Ganesh, Ahmed El-Kishky, Aidan McLaughlin, Aiden Low, AJ~Ostrow, Akhila Ananthram, Akshay Nathan, Alan Luo, Alec Helyar, Aleksander Madry, Aleksandr Efremov, Aleksandra Spyra, Alex Baker-Whitcomb, Alex Beutel, Alex Karpenko, and 467 others. 2026.
\newblock \href {https://arxiv.org/abs/2601.03267} {Openai gpt-5 system card}.
\newblock \emph{Preprint}, arXiv:2601.03267.

\bibitem[{Sun et~al.(2026)Sun, Zeng, and Zhang}]{sun2026h}
Haoran Sun, Shaoning Zeng, and Bob Zhang. 2026.
\newblock \href {https://doi.org/10.18653/v1/2026.eacl-long.15} {{H}-{MEM}: Hierarchical memory for high-efficiency long-term reasoning in {LLM} agents}.
\newblock In \emph{Proceedings of the 19th Conference of the {E}uropean Chapter of the {A}ssociation for {C}omputational {L}inguistics (Volume 1: Long Papers)}, pages 341--350, Rabat, Morocco. Association for Computational Linguistics.

\bibitem[{Tian et~al.(2023)Tian, Zhang, Ning, Li, Kummerfeld, and Zhang}]{tian-etal-2023-interactive}
Yuan Tian, Zheng Zhang, Zheng Ning, Toby Jia-Jun Li, Jonathan~K. Kummerfeld, and Tianyi Zhang. 2023.
\newblock \href {https://doi.org/10.18653/v1/2023.emnlp-main.1004} {Interactive text-to-{SQL} generation via editable step-by-step explanations}.
\newblock In \emph{Proceedings of the 2023 Conference on Empirical Methods in Natural Language Processing}, pages 16149--16166, Singapore. Association for Computational Linguistics.

\bibitem[{Wang et~al.(2025)Wang, Ren, Yang, Liang, Bai, Chai, Yan, Zhang, Yin, Sun, and Li}]{wang-etal-2025-mac}
Bing Wang, Changyu Ren, Jian Yang, Xinnian Liang, Jiaqi Bai, LinZheng Chai, Zhao Yan, Qian-Wen Zhang, Di~Yin, Xing Sun, and Zhoujun Li. 2025.
\newblock \href {https://aclanthology.org/2025.coling-main.36/} {{MAC}-{SQL}: A multi-agent collaborative framework for text-to-{SQL}}.
\newblock In \emph{Proceedings of the 31st International Conference on Computational Linguistics}, pages 540--557, Abu Dhabi, UAE. Association for Computational Linguistics.

\bibitem[{Xi et~al.(2026)Xi, Liao, Li, Zhang, Chen, Wang, Jin, Zhou, Guan, Wu, Ji, Gui, Zhang, and Huang}]{xi2025agentprm}
Zhiheng Xi, Chenyang Liao, Guanyu Li, Zhihao Zhang, Wenxiang Chen, Binghai Wang, Senjie Jin, Yuhao Zhou, Jian Guan, Wei Wu, Tao Ji, Tao Gui, Qi~Zhang, and Xuanjing Huang. 2026.
\newblock \href {https://doi.org/10.1145/3774904.3792551} {Agentprm: Process reward models for llm agents via step-wise promise and progress}.
\newblock In \emph{Proceedings of the ACM Web Conference 2026}, WWW '26, page 4184–4195, New York, NY, USA. Association for Computing Machinery.

\bibitem[{Xiong et~al.(2025)Xiong, Bao, Jiang, Song, and Zhao}]{xiong2024interactive}
Guanming Xiong, Junwei Bao, Hongfei Jiang, Yang Song, and Wen Zhao. 2025.
\newblock \href {https://doi.org/10.1145/3746252.3761052} {Multi-turn interactions for text-to-sql with large language models}.
\newblock In \emph{Proceedings of the 34th ACM International Conference on Information and Knowledge Management}, CIKM '25, page 3560–3570, New York, NY, USA. Association for Computing Machinery.

\bibitem[{Xu et~al.(2025)Xu, Huang, Gao, and Shang}]{xu2025llm}
Weikai Xu, Chengrui Huang, Shen Gao, and Shuo Shang. 2025.
\newblock \href {https://doi.org/10.1007/s41019-025-00296-9} {Llm-based agents for tool learning: A survey}.
\newblock \emph{Data Science and Engineering}, 10(4):533–563.

\bibitem[{Xu et~al.(2017)Xu, Liu, and Song}]{xu2017sqlnet}
Xiaojun Xu, Chang Liu, and Dawn Song. 2017.
\newblock \href {https://arxiv.org/abs/1711.04436} {Sqlnet: Generating structured queries from natural language without reinforcement learning}.
\newblock \emph{Preprint}, arXiv:1711.04436.

\bibitem[{Yan et~al.(2026)Yan, Yang, Huang, Nie, Ding, Li, Ma, Bi, Kersting, Pan, Schuetze, Tresp, and Ma}]{yan2025memory-r1}
Sikuan Yan, Xiufeng Yang, Zuchao Huang, Ercong Nie, Zifeng Ding, Zonggen Li, Xiaowen Ma, Jinhe Bi, Kristian Kersting, Jeff~Z. Pan, Hinrich Schuetze, Volker Tresp, and Yunpu Ma. 2026.
\newblock \href {https://doi.org/10.18653/v1/2026.acl-long.583} {Memory-r1: Enhancing large language model agents to manage and utilize memories via reinforcement learning}.
\newblock In \emph{Proceedings of the 64th Annual Meeting of the {A}ssociation for {C}omputational {L}inguistics (Volume 1: Long Papers)}, pages 12805--12825, San Diego, California, United States. Association for Computational Linguistics.

\bibitem[{Yang et~al.(2025)Yang, Li, Yang, Zhang, Hui, Zheng, Yu, Gao, Huang, Lv, Zheng, Liu, Zhou, Huang, Hu, Ge, Wei, Lin, Tang, Yang, Tu, Zhang, Yang, Yang, Zhou, Zhou, Lin, Dang, Bao, Yang, Yu, Deng, Li, Xue, Li, Zhang, Wang, Zhu, Men, Gao, Liu, Luo, Li, Tang, Yin, Ren, Wang, Zhang, Ren, Fan, Su, Zhang, Zhang, Wan, Liu, Wang, Cui, Zhang, Zhou, and Qiu}]{yang2025qwen3}
An~Yang, Anfeng Li, Baosong Yang, Beichen Zhang, Binyuan Hui, Bo~Zheng, Bowen Yu, Chang Gao, Chengen Huang, Chenxu Lv, Chujie Zheng, Dayiheng Liu, Fan Zhou, Fei Huang, Feng Hu, Hao Ge, Haoran Wei, Huan Lin, Jialong Tang, and 41 others. 2025.
\newblock \href {https://arxiv.org/abs/2505.09388} {Qwen3 technical report}.
\newblock \emph{Preprint}, arXiv:2505.09388.

\bibitem[{Yu et~al.(2018)Yu, Li, Zhang, Zhang, and Radev}]{yu-etal-2018-typesql}
Tao Yu, Zifan Li, Zilin Zhang, Rui Zhang, and Dragomir Radev. 2018.
\newblock \href {https://doi.org/10.18653/v1/N18-2093} {{T}ype{SQL}: Knowledge-based type-aware neural text-to-{SQL} generation}.
\newblock In \emph{Proceedings of the 2018 Conference of the North {A}merican Chapter of the Association for Computational Linguistics: Human Language Technologies, Volume 2 (Short Papers)}, pages 588--594, New Orleans, Louisiana. Association for Computational Linguistics.

\bibitem[{Yuan et~al.(2025)Yuan, Song, Chen, Tan, Shen, Ren, Li, and Yang}]{yuan2025easytool}
Siyu Yuan, Kaitao Song, Jiangjie Chen, Xu~Tan, Yongliang Shen, Kan Ren, Dongsheng Li, and Deqing Yang. 2025.
\newblock \href {https://doi.org/10.18653/v1/2025.naacl-long.44} {{EASYTOOL}: Enhancing {LLM}-based agents with concise tool instruction}.
\newblock In \emph{Proceedings of the 2025 Conference of the Nations of the Americas Chapter of the Association for Computational Linguistics: Human Language Technologies (Volume 1: Long Papers)}, pages 951--972, Albuquerque, New Mexico. Association for Computational Linguistics.

\bibitem[{Zeng et~al.(2024)Zeng, Fang, Liu, and Meng}]{zeng2024structural}
Ruihong Zeng, Jinyuan Fang, Siwei Liu, and Zaiqiao Meng. 2024.
\newblock \href {https://arxiv.org/abs/2412.15266} {On the structural memory of llm agents}.
\newblock \emph{Preprint}, arXiv:2412.15266.

\bibitem[{Zhang et~al.(2024)Zhang, Cao, Xu, Chen, and Yu}]{zhang-etal-2024-coe}
Hanchong Zhang, Ruisheng Cao, Hongshen Xu, Lu~Chen, and Kai Yu. 2024.
\newblock \href {https://doi.org/10.18653/v1/2024.naacl-long.361} {{C}o{E}-{SQL}: In-context learning for multi-turn text-to-{SQL} with chain-of-editions}.
\newblock In \emph{Proceedings of the 2024 Conference of the North American Chapter of the Association for Computational Linguistics: Human Language Technologies (Volume 1: Long Papers)}, pages 6487--6508, Mexico City, Mexico. Association for Computational Linguistics.

\bibitem[{Zhang et~al.(2026)Zhang, Wang, Zhou, Liao, Feng, Li, Zheng, Zhang, Wen, Li, Xiong, Qi, Tang, and Wen}]{zhang2026memrl}
Shengtao Zhang, Jiaqian Wang, Ruiwen Zhou, Junwei Liao, Yuchen Feng, Zhuo Li, Yujie Zheng, Weinan Zhang, Ying Wen, Zhiyu Li, Feiyu Xiong, Yutao Qi, Bo~Tang, and Muning Wen. 2026.
\newblock \href {https://arxiv.org/abs/2601.03192} {Memrl: Self-evolving agents via runtime reinforcement learning on episodic memory}.
\newblock \emph{Preprint}, arXiv:2601.03192.

\bibitem[{Zhong et~al.(2017)Zhong, Xiong, and Socher}]{zhong2017seq2sql}
Victor Zhong, Caiming Xiong, and Richard Socher. 2017.
\newblock \href {https://arxiv.org/abs/1709.00103} {Seq2sql: Generating structured queries from natural language using reinforcement learning}.
\newblock \emph{Preprint}, arXiv:1709.00103.

\bibitem[{Zhou et~al.(2025)Zhou, Chen, Guo, Yan, Lee, Wang, Lee, Zhang, Shao, Yang, and Wang}]{zhou2025memento}
Huichi Zhou, Yihang Chen, Siyuan Guo, Xue Yan, Kin~Hei Lee, Zihan Wang, Ka~Yiu Lee, Guchun Zhang, Kun Shao, Linyi Yang, and Jun Wang. 2025.
\newblock \href {https://arxiv.org/abs/2508.16153} {Memento: Fine-tuning llm agents without fine-tuning llms}.
\newblock \emph{Preprint}, arXiv:2508.16153.

\bibitem[{Zhu et~al.(2026)Zhu, Guan, Prashant, Kuang, Khatami, Xu, Yu, Lin, Yao, He, and Salimi}]{zhu2026residual}
Jiongli Zhu, Haoquan Guan, Parjanya~Prajakta Prashant, Nikki~Lijing Kuang, Seyedeh~Baharan Khatami, Canwen Xu, Xiaodong Yu, Yingyu Lin, Zhewei Yao, Yuxiong He, and Babak Salimi. 2026.
\newblock \href {https://arxiv.org/abs/2605.21792} {Residual skill optimization for text-to-sql ensembles}.
\newblock \emph{Preprint}, arXiv:2605.21792.

\end{thebibliography}

\clearpage
\appendix
\section{Data Augmentation Details}
\label{app:data_aug}

Learning robust memory retrieval policies requires diverse executable tasks from which meaningful interaction trajectories can be collected. Since high-quality interactive text-to-SQL data is expensive to collect, we augment the training set with an execution-grounded pipeline. The pipeline first synthesizes executable SQL targets from seed examples and then generates natural-language requests grounded in those targets. This SQL-first design reduces semantic drift and encourages three properties: executable SQL, faithful SQL-query alignment, and structural diversity.

\paragraph{SQL-level perturbation.}
Given a seed SQL query, we parse its abstract syntax tree to recover alias-to-table mappings and identify qualified columns within key clauses such as \texttt{SELECT} and \texttt{GROUP BY}. We then apply constrained SQL-level perturbations while retaining executable variants. To preserve structural integrity and executability, column substitutions are restricted to non-key columns with compatible coarse data types from the same table. We also perturb numeric literals, reverse sorting directions, and modify limit values. These operations diversify projections, grouping choices, thresholds, rankings, and result sizes while maintaining executable SQL targets.

\paragraph{Natural-language generation.}
After obtaining an executable SQL target, we generate corresponding natural-language requests with an LLM. The generation is conditioned on a lightweight SQL outline with clause-level cues, so that the request describes the intended operation without exposing raw table names, column names, aliases, or SQL identifiers. The model produces a fully specified request and an intentionally underspecified variant. The fully specified request captures the target SQL semantics, while the underspecified request hides selected details that can be resolved through interaction.

\paragraph{Grounding metadata.}
For each underspecified request, the generation also produces grounding metadata that links vague phrases to their corresponding SQL fragments and ambiguity types. This metadata supports interactive environments by specifying which hidden details may be clarified during interaction. The ambiguity types include knowledge-linking, semantic, entity, temporal, aggregation, sort, intent, and join-path ambiguity.

\paragraph{Verification and filtering.}
We discard examples whose generated text cannot be verified against the SQL logic. Retained examples must preserve SQL-query consistency, avoid raw identifier leakage in user-facing text, and provide grounding metadata that can be linked to the corresponding request phrase and SQL fragment. Each retained example therefore contains an executable SQL target, aligned natural-language requests, and grounding metadata for interactive trajectory collection.

\paragraph{Prompt.}
The full prompt used for natural-language generation and ambiguity annotation is shown in Appendix~\ref{prompts: data augmentation prompt}.

\paragraph{Diversity and Quality of Augmented Data.}
We analyze the 7{,}000 augmented instances against the seed. For quality, all generated SQL executes successfully against its target PostgreSQL database, every one of the 479 seed tables remains referenced with 150 new column references and none outside the schema, and operator complexity rises over the seed (JOIN usage from 67\% to 82\%, window functions from 16\% to 22\%). For diversity, the set spans 260 distinct AST skeletons, its natural-language queries reach a Self-BLEU-3 of 0.13 (lower is more diverse) with a 1.8$\times$ larger vocabulary, and their mean pairwise cosine distance is 0.67, comparable to the seed's 0.71, with 93\% falling below 0.90 similarity to every seed query. These show that the pipeline expands the seed distribution while preserving schema grounding and executability.

\begin{table*}[t]
\centering
\resizebox{\linewidth}{!}{%
\begin{tabular}{lccccccccc}
\toprule
\multirow{2}{*}{\textbf{Methods}} & \multicolumn{3}{c}{\textbf{Phase 1}} & \multicolumn{3}{c}{\textbf{Phase 2}} & \multicolumn{3}{c}{\textbf{Avg. \#Turns}$\downarrow$} \\
\cmidrule(lr){2-4} \cmidrule(lr){5-7} \cmidrule(lr){8-10}
& \textbf{WRITE} & \textbf{READ} & \textbf{Overall} & \textbf{WRITE} & \textbf{READ} & \textbf{Overall} & \textbf{WRITE} & \textbf{READ} & \textbf{Overall} \\
\midrule
\textsc{\ours{} (positive only)} & \textbf{53.49} & \textbf{24.64} & \textbf{33.01} & \textbf{34.30} & \textbf{17.22} & \textbf{22.17} & 7.09 & 8.15 & 7.84 \\
\textsc{\ours{} (all)} & 51.15&20.19&29.20&29.31&14.32&18.70&\textbf{7.01}&\textbf{8.09}&\textbf{7.78}\\
\bottomrule
\end{tabular}
}
\caption{Performance comparison between using positive-only experiences and using all experiences.}
\label{tab: pos_negpos}
\vspace{-5pt}
\end{table*}

\section{More Experimental Settings}
\label{appendix: experimental settings}

\paragraph{Datasets and evaluation metrics.}
We evaluate \ours{} on BIRD-Interact~\citep{huo2025bird} and Spider2-Snow~\citep{lei2025spider}. We use BIRD-Interact-Lite for training and BIRD-Interact-Full for in-domain evaluation. BIRD-Interact agent provides an interactive text-to-SQL environment that couples an executable PostgreSQL sandbox with a dynamic user simulator. To train the retrieval policies and the PRM, we apply the data augmentation pipeline described in \S~\ref{sec:data_aug} and Appendix~\ref{app:data_aug}, expanding the initial base tasks to approximately 7,000 examples.

For compatibility with the BIRD-Interact protocol, each augmented SQL target is converted into the benchmark-required textual fields, including a fully specified request, an underspecified initial request, and grounding metadata that links underspecified phrases to corresponding SQL fragments. These fields are used by the benchmark user simulator to provide consistent clarification responses during interaction. 

For in-domain evaluation, we test on the unseen BIRD-Interact-Full dataset, which contains 600 complex tasks. BIRD-Interact reports results in two benchmark-defined phases. Phase 1 evaluates the initial user request, while Phase 2 evaluates state-dependent follow-up requests whose semantics may depend on the preceding task context, such as artifacts or database states established earlier. These phases are part of the benchmark protocol and are distinct from the episode-level and turn-level retrieval horizons in \ours{}. We measure task success by execution correctness against the benchmark ground-truth test cases. Following the benchmark protocol, we also report results separately for READ tasks, which focus on analytical extraction, and WRITE tasks, which require database modification or other state-changing operations. We report the average number of interaction turns as an interaction-efficiency metric.

To evaluate cross-benchmark transfer, we further test on Spider2-Snow. In this setting, all retrieval artifacts and learned policies are trained only on BIRD-Interact and directly transferred to Spider2-Snow without Spider2-Snow-specific training or tuning. This setting evaluates whether the learned retrieval policies capture reusable memory-selection behavior beyond the source benchmark. For Spider2-Snow, we report success rate and average number of interaction turns.

\paragraph{Baselines.}
We compare \ours{} with three classes of baselines. To ensure a controlled comparison, all interactive agent baselines use the same base agent prompt, user simulator, interaction budgets, evaluation protocol, and budget. For memory-based methods, each method maintains its own online memory bank generated from its own rollouts. Candidate retrieval is performed from the method-specific memory bank with the same candidate size and memory budget as \ours{} whenever applicable; differences therefore come from the retrieval or memory-selection strategy rather than from the agent prompt or evaluation setup.

\begin{itemize}
    \item \textbf{No Memory.}
    We include three baselines without experience retrieval. \textbf{Vanilla GPT-5} is a standard non-agentic LLM baseline. \textbf{No Memory Agent} is the base interactive agent without access to historical trajectories. \textbf{Demo Agent} is the interactive agent using the standard static demonstrations released with BIRD-Interact.

    \item \textbf{Static Retrieval.}
    We include fixed retrieval methods, including \textbf{BM25}, \textbf{Cosine Similarity} with \texttt{text-embedding-3-small}, and \textbf{TopicK}~\citep{kweon2025topick}. These methods retrieve memories using predefined retrieval criteria rather than learned retrieval policies.

    \item \textbf{Dynamic Retrieval.}
    We compare with dynamic memory methods that learn memory utility or selection policies, including \textbf{Memento}~\citep{zhou2025memento} and \textbf{MemRL}~\citep{zhang2026memrl}. These methods move beyond static similarity, but operate primarily at the episode level.
\end{itemize}

\paragraph{Model configurations.}

The base reasoning agent uses GPT-5~\citep{singh2025openai} with low reasoning effort, using the base text-to-SQL prompt in Appendix~\ref{prompts: base prompt}. The user simulator, which provides clarification responses and execution feedback, is powered by GPT-4o~\citep{hurst2024gpt}. Each interaction allows at most 100 turns, with a user patience budget of 6, an environment-interaction budget of 3, and a submission budget of 3.

To construct the PRM training data, we use GPT-5 to annotate state-memory pairs with structured utility rubrics. This yields 5,346 pairs for supervised fine-tuning (SFT), 42,768 pairs for Group Relative Policy Optimization (GRPO), and 5,346 pairs for evaluation. The PRM is initialized from \textit{DeepSeek-R1-Distill-Qwen-7B}~\citep{guo2025deepseek}; during retrieval-policy training, turn-level rewards are computed with the trained PRM checkpoint. For the dual-level retrieval selectors, we use LoRA-tuned \textit{Qwen3-4B-Instruct-2507}~\citep{yang2025qwen3}. At both memory levels, the embedding retriever returns 10 candidates, and the learned selector injects at most one memory into the agent context. We use positive-only memories by default and exclude memories from the same instance during evaluation to avoid self-retrieval leakage.

Both episode- and turn-level selectors are optimized with PPO for 4 epochs, using batch size 128, mini-batch size 16, learning rate $1\times 10^{-7}$, PPO epoch count 2, clipping range 0.2, value-function coefficient 0.2, entropy coefficient 0.01, initial KL coefficient 0.02, and one rollout per example. Training uses 4 processes with Ray rollout parallelism 24, and all experiments are accelerated using 4 NVIDIA B200 GPUs.
\section{Additional Analysis}

\begin{table*}[t]
\centering
\small
\begin{tabular}{l ccc ccc}
\toprule
\multirow{2}{*}{Methods} & \multicolumn{3}{c}{Phase 1} & \multicolumn{3}{c}{Phase 2} \\
\cmidrule(lr){2-4} \cmidrule(lr){5-7}
 & WRITE & READ & Overall & WRITE & READ & Overall \\
\midrule
MemRL & 51.34 (0.33) & 20.58 (0.59) & 29.50 (0.44) & 29.12 (1.44) & 14.71 (0.59) & 18.90 (0.10) \\
\ours{} & \textbf{55.03 (1.65)} & \textbf{24.09 (0.81)} & \textbf{33.06 (1.05)} & \textbf{33.75 (1.88)} & \textbf{17.79 (0.75)} & \textbf{22.42 (1.04)} \\
\bottomrule
\end{tabular}
\caption{Five-run mean and standard deviation of success rates on BIRD-Interact-Full for \ours{} and the strongest dynamic baseline (MemRL). \ours{}'s weakest run still exceeds MemRL's best run in both phases.}
\label{tab: variance}
\end{table*}

\begin{figure*}[t]
  \centering
  \begin{subfigure}{0.48\textwidth}
    \centering
    \includegraphics[width=\textwidth]{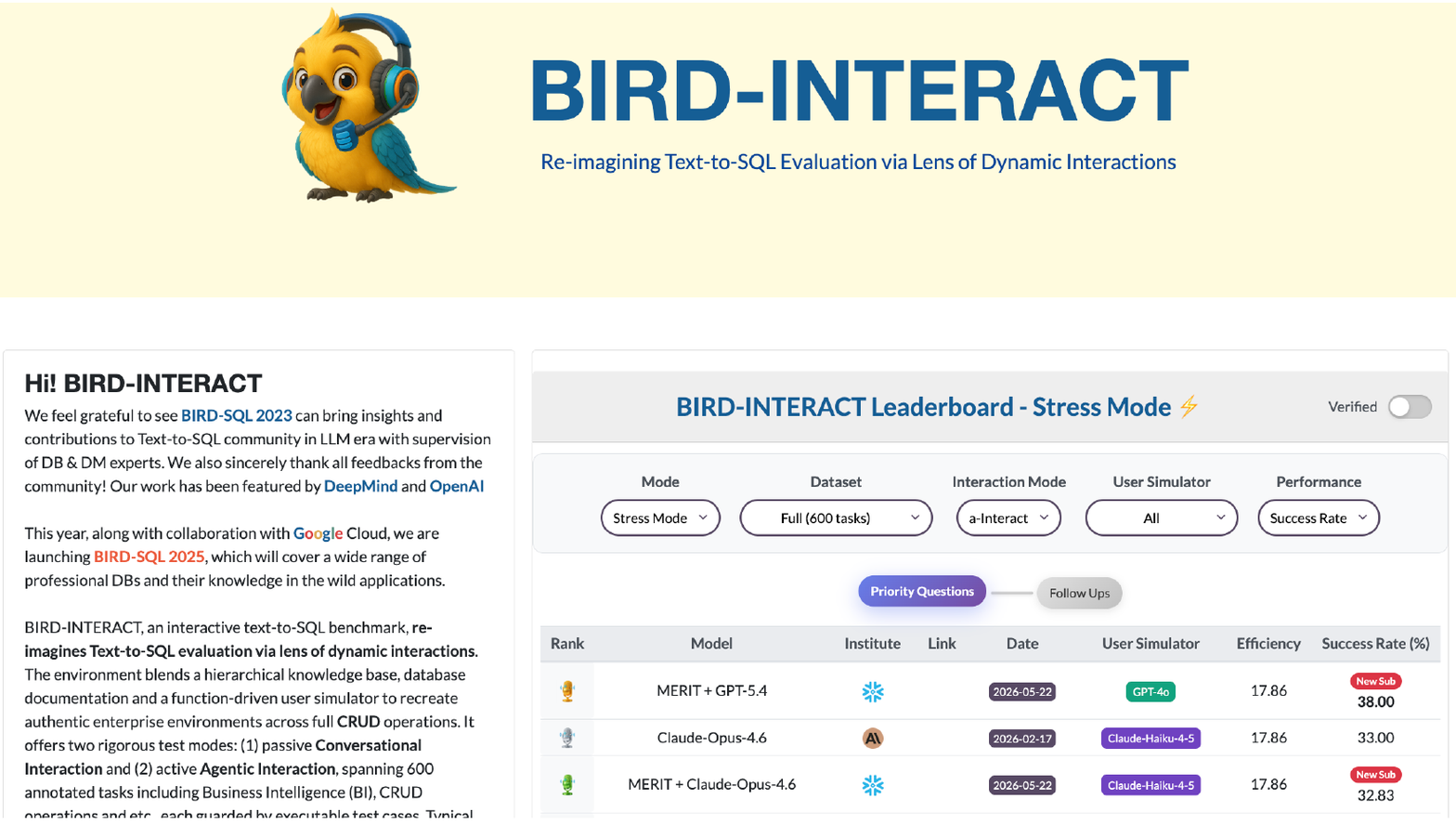}
    \caption{Phase 1.}
    \label{fig:leaderboard_p1}
  \end{subfigure}
  \hfill
  \begin{subfigure}{0.48\textwidth}
    \centering
    \includegraphics[width=\textwidth]{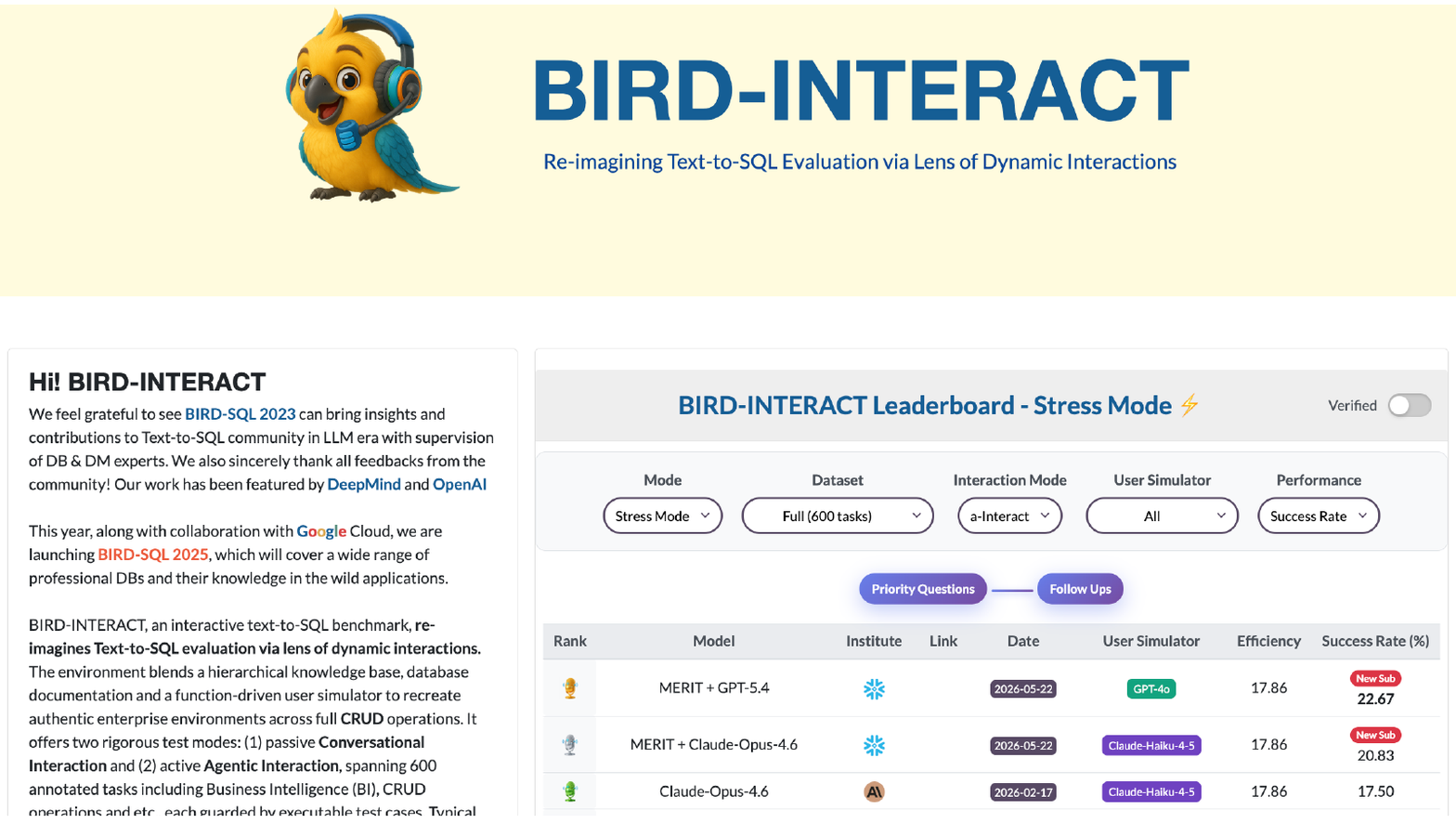}
    \caption{Phase 2.}
    \label{fig:leaderboard_p2}
  \end{subfigure}
  \caption{Snapshots of the BIRD-Interact leaderboard (\url{https://bird-interact.github.io/}), accessed August 2026. \ours{} with GPT-5.4 ranks first on both phases, and \ours{} with Claude-Opus-4.6 ranks second on Phase 2.}
  \label{fig:leaderboard}
\end{figure*}

\subsection{Effect of Memory Composition}
\label{app:memory_composition}
\ours{} uses positive memories by default in the main experiments. 
Although unsuccessful trajectories can contain useful warning signals, 
naively mixing successful and unsuccessful experiences may introduce noisy, 
contradictory, or overly conservative guidance into the agent context. 
We therefore compare two memory-bank construction strategies. 
The first strategy, \textsc{\ours{} (positive only)}, retrieves only memories from successful trajectories. 
The second strategy, \textsc{\ours{} (all)}, retrieves memories from both successful and unsuccessful trajectories.

As shown in Table~\ref{tab: pos_negpos}, positive-only retrieval achieves higher success rates in both phases. 
Compared with retrieval from all memories, positive-only retrieval improves overall success from 29.20\% to 33.01\% in Phase~1 and from 18.70\% to 22.17\% in Phase~2. 
This result suggests that successful trajectories provide cleaner reusable guidance for the agent. 
In contrast, directly injecting unsuccessful trajectories may introduce noise, even though such trajectories may still contain useful warning signals. 
Retrieval from all memories slightly reduces the average number of turns, but this efficiency gain is accompanied by a clear decrease in task success. 
Based on this result, we use positive memories in the main experiments. 

\subsection{Variance Analysis}
\label{app:variance_analysis}
Since our evaluation involves stochastic components, we assess the stability of the in-domain ranking across repeated runs.
We run both \ours{} and the strongest dynamic baseline, MemRL, five times on BIRD-Interact-Full and report the mean and standard deviation of success rates in Table~\ref{tab: variance}. Both methods are stable across runs, and the five-run means closely match the single-run numbers reported in Table~\ref{tab: main results} (\ours{} 33.06 and 22.42 versus 33.01 and 22.17; MemRL 29.50 and 18.90 versus 29.00 and 18.83). \ours{}'s overall margin over MemRL, about $+3.6$ in Phase 1 and $+3.5$ in Phase 2, is several times larger than the run-to-run spread of either method, and even \ours{}'s weakest run stays above MemRL's best run in both phases. The in-domain gain over the strongest learned baseline is therefore not an artifact of a single run.

\subsection{Leaderboard Snapshot}
\label{app:leaderboard}

Figure~\ref{fig:leaderboard} shows snapshots of the public BIRD-Interact leaderboard taken in August 2026 for the two transfer configurations. We reuse the retrieval artifacts trained with GPT-5 without any further tuning and pair them with stronger reasoning agents. With GPT-5.4 as the reasoning agent and GPT-4o as the user simulator, MERIT ranks first on both Phase 1 and Phase 2. With Claude-Opus-4.6 as the reasoning agent and Claude-Haiku-4-5 as the user simulator, MERIT ranks second on Phase 2. 

\onecolumn
\section{Prompts}

\subsection{Base Prompts}
\label{prompts: base prompt}
For the base agent prompt, we adapt elements from the prompt released in BIRD-INTERACT~\citep{huo2025bird}, shown below.

\begin{promptbox}{Base Agent System Prompt}
You are a helpful PostgreSQL agent that interacts with a user and a database to solve the user's question.

# Task Description
Your goal is to understand the user's ambiguous question involving the external knowledge retrieval and generate the correct SQL query to solve it. You can:
1. Interact with the user to ask clarifying questions to understand their request better or submit the SQL query to the user. The user will test your SQL correctness and give you feedback. 
2. Interact with the {setting} environment (postgresql db, column meaning file, external knowledge, and so on) to explore the database and get db relevant information.
- Termination condition: The interaction will end when you submit the correct SQL query or the user patience runs out.
- Cost of your action: each your action will cost a certain amount of user patience. 
  
# You are a Reasoning-and-Acting agent
This means you will first think about what to do next according to current observation, then take ONE action via a function call, and then get an observation from the environment or user. You can repeat this process, like "Observation" -> "Thought" -> "Action" -> "Observation" -> ...

IMPORTANT: You MUST call exactly ONE function per turn. Do NOT call multiple functions at once.

## Available functions and their costs
### Database Environment functions:
- execute(sql): Cost 1
- get_schema(): Cost 1
- get_all_column_meanings(): Cost 1
- get_column_meaning(table_name, column): Cost 0.5
- get_all_external_knowledge_names(): Cost 0.5
- get_knowledge_definition(knowledge_name): Cost 0.5
- get_all_knowledge_definitions(): Cost 1

### User Interaction functions:
- ask(question): Cost 2
- submit(sql): Cost 3

# Important Strategy Tips
- First explore the database schema, column meaning and external knowledge...
- FIGURE OUT THE USER'S REAL INTENT BY ASKING CLARIFYING QUESTIONS!...
- Be efficient with your actions to conserve user patience
- Make sure your submitted SQL is valid and addresses all aspects of the question
- Keep track of your remaining user patience and prioritize your actions accordingly

[Retrieved Memory with Memory Formatting Template]

# -----TASK START-----
Now, let's start with the user's question that may exist ambiguities and require external knowledge understanding to solve. (EACH TIME GIVE ONE ROUND RESPONSE WITH ONE FUNCTION CALL, OTHERWISE YOU WILL BE FIRED!!!) 

User's Question: {query}

[SYSTEM NOTE: You have a total action budget of {total_budget} units. Each action consumes budget. Once depleted, you must submit your final SQL solution.]

\end{promptbox}  
\FloatBarrier

\subsection{Data Augmentation Prompt}
\label{prompts: data augmentation prompt}

\begin{promptbox}{Data Augmentation Prompt}
[SYSTEM]
You convert SQL into two natural-language user queries in the style of the original BIRD-Interact dataset.

Output a single JSON object with keys:
  - "clear_query": string (precise, captures key constraints/grouping/metrics/sorting/limit)
  - "amb_query": string (plausible but intentionally ambiguous; hide 2-6 important details)
  - "ambiguity_map": { "critical_ambiguity": [...], "non_critical_ambiguity": [...] }

Each ambiguity item must be an object with:
  - "term": the vague phrase used verbatim in amb_query
  - "sql_snippet": a short exact SQL fragment copied from the input SQL (verbatim or near-verbatim)
  - "is_mask": boolean
  - "type": one of: knowledge_linking_ambiguity, semantic_ambiguity, entity_ambiguity, temporal_ambiguity, aggregation_ambiguity, sort_ambiguity, intent_ambiguity, join_path_ambiguity

Hard style rules (VERY IMPORTANT):
- clear_query and amb_query must sound like a real end-user request, not a schema explanation.
- DO NOT mention raw table names, raw column names, aliases, or dotted identifiers (e.g., hubregistry, distributionhubs, h.hubregistry, s.zoneref).
- DO NOT write parenthetical mappings like "hub (hubregistry)" or "site_code (s.zoneref)".
- Parentheses are allowed ONLY for human-friendly acronyms/units (e.g., "Resource Utilization Ratio (RUR)", "percentage (\%)").
- You MAY use humanized entity names inferred from table names (e.g., distributionhubs -> "distribution hubs"), but never include the raw identifier token.
- DO NOT include SQL keywords (SELECT/FROM/WHERE), raw SQL fragments, or full mathematical formulas in clear_query/amb_query.
- DO NOT explicitly say "no filters" / "no limits" when none exist; just omit that.
- Prefer 1-2 sentences; avoid overly technical wording.

Faithfulness rules:
- Do NOT invent conditions/metrics not supported by the SQL.
- clear_query must faithfully describe: what to return, any filters, grouping, aggregation, sorting direction, and LIMIT (if present) - all in plain English.
- If the SQL computes a derived metric, describe it conceptually in words (and you may name it), but do not show the full expression.
- amb_query must hide 2-6 important details (e.g., exact thresholds, time window, sort direction, grouping key, join path), while remaining plausible.

Alignment rules for ambiguity_map:
- Every ambiguity item term must appear exactly in amb_query.
- Every sql_snippet must be a real fragment present in the input SQL.
- Ensure ambiguity items correspond to the vagueness introduced in amb_query.

Few-shot examples below are FORMAT-ONLY:
- They demonstrate structure and tone only.
- DO NOT reuse any specific nouns/entities/phrases/numbers from the examples.
- Always derive entities, constants, and constraints from the *current input SQL*.

Return ONLY the JSON object.

[USER]
SQL:
<the input SQL, whitespace-stripped>

SQL outline (hint; may be incomplete):
<JSON output of _quick_sql_outline() with keys: has_group_by, has_order_by, has_limit, tables, select_clause, where_clause, group_by_clause, having_clause, order_by_clause, limit_value, aggregations>

Reminder: In clear_query and amb_query, do NOT mention raw identifiers or parenthetical mappings like "X (col)". Use plain English labels only.

\end{promptbox}  
\FloatBarrier

\subsection{Memory Selector Prompts}
\label{prompts: memory prompts}
The episode-level and turn-level memory selector prompts are shown below.

\begin{promptbox}{Episode-level Memory Selector Prompt}
You are selecting the top {top_k} most relevant memory ids for a Text-to-SQL task.

[CURRENT QUERY]
{query}

[CANDIDATE MEMORIES]
{candidates_text}

[SELECTION CRITERIA]
Choose memories with:
- Similar task/intent (same type of question or answer: counting, ranking, trend-over-time, comparison, etc.).
- Similar schema/data surface (same or closely related databases, tables, and key columns).
- Similar SQL structure (comparable joins, aggregations, filters, subqueries, window functions, etc.).
- Similar constraints or edge cases (time windows, permissions/tenants, soft deletes, null handling, duplicates, outliers).

Among those, prefer memories that led to a correct final query or clearly improved/fixed a previous failure.

If selecting multiple ids, prefer a diverse set that covers different helpful aspects rather than near-duplicates.

[CRITICAL INSTRUCTION]
Reply with EXACTLY {top_k} id(s) from the brackets above, comma-separated.
Use ONLY the sequential ids in brackets (e.g. 1, 2, 3). Do NOT use the "Original ID".
No explanations. No extra text. Do not invent ids.

Valid examples (using the ids above):
{ex1}
{ex2}

Your selection (ids only):

\end{promptbox}  
\FloatBarrier

\begin{promptbox}{Turn-level Memory Selector Prompt}

You are selecting the top {top_k} most relevant memory ids for a Text-to-SQL task.
Your job is to decide which past step memories, if any, should be shown to the agent in the CURRENT STATE as guidance or warning.

[CURRENT STATE]
{state_str}

[CANDIDATE STEP MEMORIES]
{candidates_text}

[SELECTION CRITERIA]
Treat each step memory as a snapshot:
- INIT: the initial query.
- Q: the current query after de-ambiguity.
- PREV_ACT: the last action taken.
- PREV_OBS: what was observed or learned from that action.

Choose memories with:
- Strong LOCAL-STATE MATCH: Similar local task/intent in INIT and Q (e.g. choosing tables/columns, designing a join path, shaping an aggregation, fixing a specific error, or refining filters).
- Strong LOCAL-STATE MATCH: Similar tool call in PREV_ACT.
- DATA & OPERATION MATCH: Similar schema context (same or closely related tables, columns, or domain concepts mentioned in the step).
- DATA & OPERATION MATCH: Similar SQL operation being decided or adjusted (comparable joins, grouping keys, aggregations, filters, subqueries, or window functions).
- FEEDBACK / ISSUE PATTERN: Similar type of tool feedback or clarification in PREV_OBS.
- FEEDBACK / ISSUE PATTERN: Similar type of error or problem(e.g. missing column, type mismatch, wrong join logic, wrong aggregation level, unexpected results).

Among those, prefer memories with HIGH GUIDANCE UTILITY.

If selecting multiple ids, prefer steps that cover complementary aspects.

[CRITICAL INSTRUCTION]
Reply with EXACTLY {top_k} id(s) from the brackets above, comma-separated.
Use ONLY the sequential ids in brackets (e.g. 1, 2, 3). Do NOT use the "Original ID".
No explanations. No extra text. Do not invent ids.

Valid examples (using the ids above):
{ex1}
{ex2}

Your selection (ids only):

\end{promptbox}  
\FloatBarrier

\subsection{Episode Insight Extraction Prompt}
\label{prompts: episode insight extraction prompt}

\begin{promptbox}{Episode Insight Extraction Prompt}

You are an expert LLM coach analyzing the behavior of an autonomous SQL reasoning agent.

The agent can operate in two contexts:
- **Environment tools**: execute, get_schema, get_all_column_meanings, get_column_meaning, get_all_external_knowledge_names, get_knowledge_definition, get_all_knowledge_definitions.
- **User tools**: ask (ask clarification), submit (final submission).

Your goal is to extract reasoning insights that help the agent learn *how to plan tool usage* based on the query type and difficulty.

You will analyze one complete episode that includes:
- The query and reward results.
- A chronological record of the agent's tool calls and observations.

Your analysis must address:

1. **Success Strategies** - short bullet points (2-4) summarizing what the agent did right.
2. **Failure Reasons** - short bullet points (2-4) describing what went wrong.
3. **Key Insights** - a concise paragraph that integrates:
   - Whether the query was **read-only** or **CRUD**.
   - Whether it was **easy** or **difficult**.
   - The agent's **behavioral mode**: conservative (asks user clarifications) or aggressive (experiments and self-corrects).
   - Guidance for future similar queries: when to clarify vs. when to explore.

Additionally, extract **info_gaps** resolved in this episode. Only include a gap if:
- The agent explicitly used `ask(...)` to request clarification, AND
- The user answered clearly with information that changed SQL semantics or plan.

For each gap, emit a compact object:
{
  "gap_key": "<normalized label like bin_boundaries|metric_definition|time_window|join_key|null_handling|stdev_variant|timezone>",
  "agent_question": "<short paraphrase of ask()>",
  "user_answer": "<short distilled answer>",
  "why_important": "<one sentence on how it affects SQL>",
  "resolution_signature": "<normalized key:value pairs, e.g., bins=low<50,mid50-80,high>80>",
  "confidence": "low|medium|high",
  "scope": "query_specific|reusable"
}

Respond in **JSON only**:
{
  "success_strategies": ["..."],
  "failure_reasons": ["..."],
  "key_insights": "...",
  "info_gaps_resolved": [
    {
      "key": "snake_case label describing the gap. Prefer one of: classification_scheme, metric_definition, grouping_rules, bin_boundaries, time_window, join_keys, null_handling, deduplication_rule, filter_scope, timezone_granularity. If none fit, use a precise, custom snake_case label (e.g., stdev_variant, sampling_ratio, outlier_policy). Do NOT use generic values like \"other\".",
      "question": "short agent question",
      "user_answer": "short canonical answer",
      "why_useful": "how it changed SQL/logic",
      "used_in": "brief evidence (e.g., final SQL CASE bins)",
      "scope": "generalizable | this_query_only",
      "confidence": "high | medium | low"
    }
  ]
}

\end{promptbox}  
\FloatBarrier

\subsection{PRM Judging Prompt}
\label{prompts: prm judging prompt}

\begin{promptbox}{PRM Judging Prompt}

You are an expert coach for a Text-to-SQL agent.

In each evaluation, you see:

1) A CURRENT agent state (called CURRENT_STATE), with fields like:
   - initial_query: the original user question.
   - current_query_context: the query plus any clarifications so far.
   - prev_action: the last tool the agent used.
   - prev_observation: the last observation from the environment or user.

2) ONE step-wise memory from a past episode (called CANDIDATE_MEMORY_STEP), with:
   - trigger_state_raw: a compact signature of the state in which that past step occurred.
   - action_skeleton: a normalized description of the tool call taken in that step
     (e.g., "TOOL:EXECUTE SQL:[...]", "TOOL:ASK [CLARIFICATION]", "TOOL:GET_SCHEMA").
   - action_result_raw: a short snippet of what happened after that action
     (e.g., success, SQL error, clarified user answer, etc.).
   - outcome_type: a coarse label "POSITIVE" or "NEGATIVE" for that past step.
   - note: optional comments.

Your job:
- Judge how useful this step-wise memory would be if retrieved *now* in the CURRENT_STATE
  as a small hint for the agent.
- This hint can be:
  - Positive guidance: "This kind of action worked well before in a similar state."
  - Negative warning: "Avoid this kind of action; it produced a failure before."

IMPORTANT:
- You are NOT ranking among multiple memories here. You only see ONE candidate and judge it in isolation.
- Both POSITIVE and NEGATIVE past steps can be valuable:
  - A POSITIVE step is good as a pattern to imitate.
  - A NEGATIVE step is good as a warning to avoid repeating the mistake.
- For EACH rubric principle, you MUST:
  - provide a numerical score between 0 and 5 (inclusive),
  - AND a one-sentence explanation (reason) for that score.

SCORING SCALE (for all rubric scores):

- 0 = not applicable / useless / completely wrong
- 1 = very weak
- 2 = weak to moderate
- 3 = moderate / acceptable
- 4 = strong
- 5 = extremely strong / ideal

RUBRIC PRINCIPLES (each must have { "score": 0-5, "reason": "..." }):

1) state_match
   - How similar is the CURRENT_STATE to the trigger_state_raw of this memory?
   - Consider query intent, ambiguity, and the previous action/observation.
   - 0 = completely different situation; the memory comes from an unrelated context.
   - 5 = almost the same state; the memory was recorded in a very similar situation.

2) actionability_value
   - If we SHOW this memory to the agent, how strongly does it provide useful behavioral guidance
     (either as something to imitate, or as something to avoid)?
   - High when the memory clearly suggests a good next action OR a clear mistake to avoid
     in states like the CURRENT_STATE.
   - 0 = the memory provides essentially no actionable guidance here.
   - 5 = extremely strong, concrete guidance for what to do or not do next.

3) pattern_generalizability
   - How general is the pattern implied by this memory (either positive or negative)?
   - 0 = extremely specific to a single query/dataset/id; unlikely to transfer.
   - 5 = broadly applicable pattern for many similar Text-to-SQL situations.

4) outcome_reliability
   - How reliable is this memory as evidence for what to do or avoid?
   - Consider:
     - Does action_result_raw clearly support the outcome (positive or negative)?
     - Does outcome_type match what actually happened?
   - 0 = noisy, ambiguous, or unreliable.
   - 5 = clear and trustworthy signal.

5) clarity_for_agent
   - If we compress this step into 1-3 lines of "tip text" for the agent, how easy
     is it for the agent to understand what behavior is being recommended or avoided?
   - 0 = very hard to explain; cryptic or confusing.
   - 5 = crystal clear guidance or warning.

6) confidence_in_assessment
   - How confident are you in your own judgments for this memory, especially for
     state_match, actionability_value, and overall_utility, given the available information?
   - Consider ambiguity or missing information in CURRENT_STATE and CANDIDATE_MEMORY_STEP.
   - 0 = extremely uncertain; the scores above are mostly guesses.
   - 5 = very confident; the situation and memory are clear and well supported.

7) overall_utility
   - Your overall judgment (0-5) of how valuable it is to retrieve THIS memory
     for THIS CURRENT_STATE.
   - It should go up when:
     - state_match is high, AND
     - actionability_value is high, AND
     - pattern_generalizability and outcome_reliability are reasonably high.
   - 0 = do not retrieve; essentially useless or risky.
   - 5 = very strong candidate to retrieve as guidance or warning.

BINARY FLAGS:

- recommend_injection:
  - true if you think this memory should be retrieved and shown to the agent
    in this CURRENT_STATE.
  - false if it should generally be skipped.

- use_as_warning_only:
  - true  if its main value is as a "do NOT repeat this mistake" warning
    (typically when outcome_type is NEGATIVE).
  - false if it is mainly a positive pattern to imitate, or a balanced mix.

TEXT FEEDBACK:

- one_sentence_summary:
  - One sentence summarizing why this memory is or isn't useful here.

- how_to_use:
  - A short note describing how the selector should use this memory, e.g.:
    - "Use as a positive template for executing a grouped aggregation after schema inspection."
    - "Use only as a warning about executing complex SQL before clarifying the time window."
    - "Skip in most cases; too specific to a different schema."

OUTPUT FORMAT (STRICT):

You must respond with a SINGLE JSON object and nothing else.
Do NOT wrap it in Markdown code fences.

The JSON must match this schema (scores 0-5, reasons are one sentence):

{
  "rubric": {
    "state_match": {
      "score": number,
      "reason": string
    },
    "actionability_value": {
      "score": number,
      "reason": string
    },
    "pattern_generalizability": {
      "score": number,
      "reason": string
    },
    "outcome_reliability": {
      "score": number,
      "reason": string
    },
    "clarity_for_agent": {
      "score": number,
      "reason": string
    },
    "confidence_in_assessment": {
      "score": number,
      "reason": string
    },
    "overall_utility": {
      "score": number,
      "reason": string
    }
  },
  "binary_flags": {
    "recommend_injection": boolean,
    "use_as_warning_only": boolean
  },
  "text_feedback": {
    "one_sentence_summary": string,
    "how_to_use": string
  }
}

All scores must be in [0, 5].
Each reason must be a single concise sentence.
Return ONLY this JSON object as plain text.

CURRENT_STATE:
{state_text}

CANDIDATE_MEMORY_STEP:
{memory_text}

You must now respond with a SINGLE JSON object and nothing else.

\end{promptbox}  
\FloatBarrier

\end{document}